\documentclass[10pt,twocolumn,letterpaper]{article}

\usepackage[pagenumbers]{cvpr}

\usepackage{amsmath}
\usepackage{amssymb}
\usepackage{algorithm}
\usepackage{algpseudocode}
\usepackage{booktabs}
\usepackage{graphicx}
\usepackage{float}
\usepackage{multirow}
\usepackage{xcolor}
\usepackage{caption}
\usepackage{subcaption}

\usepackage{colortbl}
\usepackage{booktabs}
\usepackage{array}

\usepackage{booktabs}
\usepackage{array}
\usepackage{colortbl}

\definecolor{cvprHeatBlue}{HTML}{356CA5}
\definecolor{cvprOverallGray}{HTML}{ECEEF1}
\definecolor{cvprOursGray}{HTML}{F1F3F5}

\providecommand{\cvprTableBest}[1]{}
\renewcommand{\cvprTableBest}[1]{\textbf{#1}}

\providecommand{\cvprTableSecond}[1]{}
\renewcommand{\cvprTableSecond}[1]{\underline{#1}}

\providecommand{\cvprTableHeat}[2]{}
\renewcommand{\cvprTableHeat}[2]{%
  \cellcolor{cvprHeatBlue!#1!white}#2%
}

\definecolor{cvprHeatGreen}{HTML}{238B57}

\renewcommand{\cvprTableHeat}[2]{%
  \cellcolor{cvprHeatGreen!#1!white}#2%
}

\definecolor{cvprblue}{rgb}{0.21,0.49,0.74}
\usepackage[pagebackref,breaklinks,colorlinks,allcolors=cvprblue]{hyperref}

\usepackage{colortbl}

\newcommand{\gup}[1]{\textcolor{green!55!black}{#1}}
\newcommand{\gdn}[1]{\textcolor{red!75!black}{#1}}
\IfFileExists{fontawesome5.sty}{\usepackage{fontawesome5}}{}

\def\confName{CVPR}
\def\confYear{2027}

\title{SEE: Unlocking the Hidden Perception of Frozen VLMs}

\author{Alin-Ionut Popa \qquad Tuo Zhang \qquad Dimitrios Dimitriadis\\ Amazon\\ {\tt\small \{popaaln, tuozhang, dbdim\}@amazon.com}}

\renewcommand{\topfraction}{0.85}\renewcommand{\textfraction}{0.1}

\makeatletter\renewcommand\paragraph{\@startsection{paragraph}{4}{\z@}{1.2ex \@plus 0.4ex \@minus 0.2ex}{-1em}{\normalfont\normalsize\bfseries}}\makeatother
\begin{document}
\twocolumn[{%
\renewcommand\twocolumn[1][]{#1}%
\maketitle
\begin{center}
  \centering
  \includegraphics[width=0.93\textwidth]{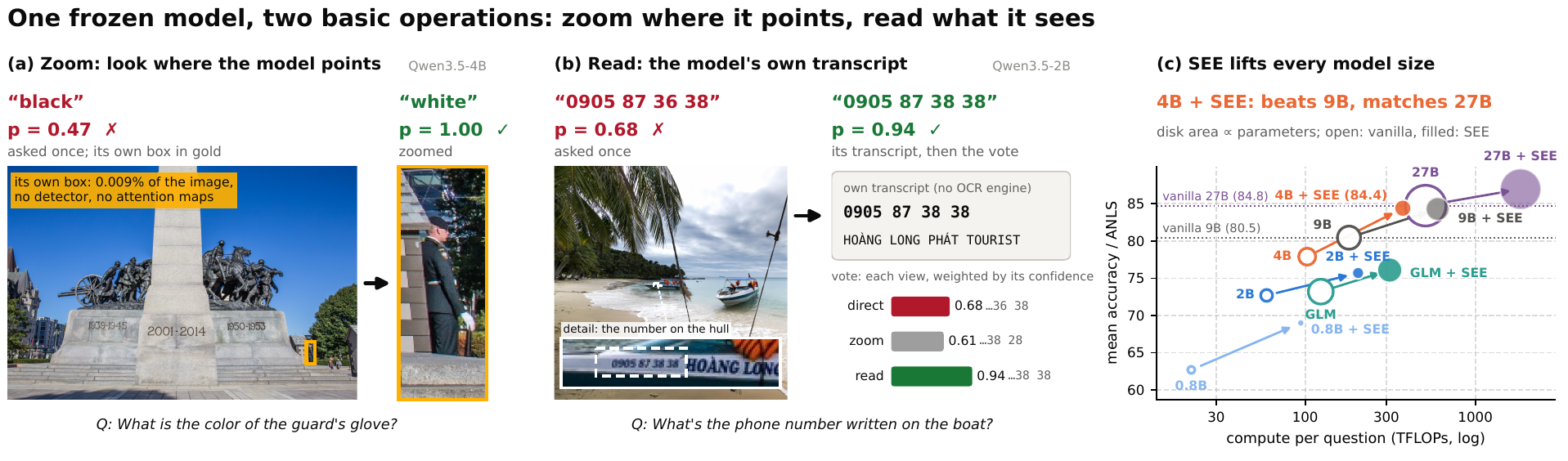}
  \captionof{figure}{\textbf{SEE lets a frozen VLM look again where it points and read what it sees.} (a)~\texttt{Zoom}: asked once,
Qwen3.5-4B gets the color of a glove covering $0.009\%$ of the image wrong, yet its own box lands on the glove;
answering again on that zoom gets it right, with no detector or attention maps. (b)~\texttt{Read}: the direct answer
and the zoom misread a phone number (dashed box and inset: its location, for the reader), while the model's own
transcript reads it, so the confidence-weighted vote picks the reading view. (c)~With SEE, Qwen3.5-4B beats the direct answer of Qwen3.5-9B
and matches that of the $6{\times}$ larger Qwen3.5-27B, at lower compute (TFLOPs per question; App.~F).}
  \label{fig:teaser}
\end{center}
}]

\begin{abstract}
Vision–language models (VLMs) grasp the overall content of an image but often miss the fine details a question hinges on, such as a small object in a large photo or a number in a dense infographic. Yet even when their answer is wrong, the region they point to often contains the evidence.
We introduce Self-Elicited Evidence (SEE), a training-free method that lets a frozen VLM look again where it points. When the initial answer has low confidence, SEE prompts the model to propose regions where the details may appear, answers the question on these enlarged crops, and aggregates the answers with the original prediction through confidence-weighted voting. For text-rich images, the model also answers with its own transcript of the image. SEE needs no detector or OCR engine.
Across five backbones, SEE improves every small-object benchmark by 2.3–13.1 points and outperforms five training-free baselines on small objects. With SEE, Qwen3.5-4B surpasses the direct answer of Qwen3.5-9B and matches that of the $6{\times}$ larger Qwen3.5-27B.

\end{abstract}

\section{Introduction}
\label{sec:intro}
Vision–language models (VLMs)~\cite{qwen2025,glm41v} can capture the overall content of an image yet miss fine details needed to answer specific questions~\cite{vstar2024,mmvp2024,vicrop2025}. A model may recognize a street scene but misjudge the relation between two small objects~\cite{vstar2024,hrbench2024}, or understand a document while misreading a crucial number~\cite{infographicvqa2021,textmonkey2024}. These errors expose a gap between broad image understanding and question-specific visual evidence. Can a frozen VLM recover useful evidence for a question it initially answers incorrectly?

An incorrect answer does not necessarily imply an inability to locate the relevant evidence. We find that a frozen VLM can identify the queried region even when it fails to answer the question~\cite{vicrop2025}, suggesting that its localization predictions~\cite{kosmos2} may guide further inspection. Figure~\ref{fig:detail-inspection} illustrates this opportunity: the model gives an incorrect direct answer but selects a relevant region and answers correctly from the crop. The annotated evidence accounts for 1.99\% of image attention~\cite{pai2024eccv,visualsink2025} in the full-image pass and 11.33\% in the crop pass. These observations motivate using the model’s own predictions to obtain additional evidence before making a final decision.

Additional visual inspection offers a way to recover details missed in an initial answer. Existing approaches learn to search~\cite{vstar2024} or zoom while reasoning~\cite{deepeyes2025,chainoffocus2025,minio32025,pixelreasoner2025}, or select regions at inference time using attention maps~\cite{vicrop2025,hide2025}, external detectors~\cite{dyfo2025}, and patch-tree search~\cite{zoomeye2024}. Our observation suggests a simpler source of guidance: the frozen model's own localization predictions. Predicted-box zooming already exploits this capability by asking the model to select a region and answer again~\cite{zoomrefine2025}.
These approaches demonstrate the value of further visual inspection. Our observation suggests that a model's initial answer may understate its ability to extract relevant evidence: even after answering incorrectly, the same model can still identify where to look. This raises a broader question: to what extent can fine-detail errors be recovered by eliciting and using evidence that is already accessible to the frozen model?

To investigate this question, we introduce Self-Elicited Evidence (SEE), a training-free method that uses a frozen VLM to generate and evaluate complementary evidence, as illustrated in Figure~\ref{fig:teaser}. SEE samples several self-proposed regions and answers the question on their enlarged crops. It combines these answers with the original prediction through confidence-weighted voting, rather than allowing a single crop to determine the result. For text-rich inputs, SEE also obtains the model’s own transcription and answers with both the transcript and the full image. This reading view provides an alternative when local crops omit relevant text or context. A confidence gate skips additional inspection for high-confidence direct answers. SEE uses model outputs and their token probabilities, without an external detector, OCR engine, or access to attention maps.

We evaluate on V*Bench~\cite{vstar2024}, HR-Bench 4K and 8K~\cite{hrbench2024}, and InfographicVQA~\cite{infographicvqa2021} with four Qwen3.5~\cite{qwen2025} sizes and GLM-4.1V~\cite{glm41v}. Our contributions are threefold. (a)~We show that a frozen VLM often localizes the evidence it fails to use in its direct answer~\cite{vicrop2025} (Sec.~\ref{sec:elicit}). (b)~We propose SEE, a training-free method that, when unsure, zooms into self-proposed regions and adds the model's own transcript to the full image, then lets every view, the direct answer included, vote with its confidence (Sec.~\ref{sec:method}). (c)~On three small-object benchmarks, SEE improves every model--benchmark pair, raises mean accuracy from 72.4\% to 78.2\%, and outperforms all five training-free baselines; random-crop and fixed-grid controls fall well short (Sec.~\ref{sec:experiments}).

\section{Related Work}
\label{sec:related}

\begin{table}[t]
\centering
\caption{\textbf{How SEE differs from the training-free methods we compare.} Only SEE reads: it adds the model's own transcript
of the image as a view. \emph{Needs}: what a method uses beyond the model's text output; \emph{New views}: what it
adds to the whole image; \emph{Decides}: what picks the final answer; \emph{Time}: per question, as a multiple of the
direct answer (Qwen3.5-4B, Sec.~\ref{sec:cost}). Self-consistency runs with SEE's gate and vote.}
\setlength{\tabcolsep}{2pt}
\footnotesize
\begin{tabular}{@{}l@{\hspace{3pt}}l@{\hspace{3pt}}l@{\hspace{2pt}}c@{\hspace{3pt}}l@{\hspace{2pt}}r@{}}
\toprule
\textbf{Method} & \textbf{Needs} & \textbf{New views} & \textbf{Reads} & \textbf{Decides} & \textbf{Time} \\
\midrule
Self-cons.~\cite{selfconsistency2022} & probs. & chains & -- & conf. vote & $5.0\times$ \\
Zoom-Refine~\cite{zoomrefine2025} & -- & one box & -- & 2nd pass & $2.3\times$ \\
ZoomEye~\cite{zoomeye2024} & probs. & tree search & -- & chosen patches & $127\times$ \\
HiDe~\cite{hide2025} & attention & packed regions & -- & one pass & $3.1\times$ \\
DyFo~\cite{dyfo2025} & detector & crops & -- & reward vote & $20\times$ \\
\midrule
\textbf{SEE} & probs. & own boxes, text & \checkmark & conf. vote & $5.9\times$ \\
\bottomrule
\end{tabular}
\label{tab:position}
\end{table}

\paragraph{Visual evidence acquisition.}
V*~\cite{vstar2024} trains a visual search model, while recent approaches learn to zoom during reasoning~\cite{deepeyes2025,chainoffocus2025,minio32025,pixelreasoner2025}. These methods require additional training to acquire an inspection policy. Training-free alternatives avoid this cost but introduce other dependencies: ViCrop~\cite{vicrop2025} and HiDe~\cite{hide2025} require access to attention maps, DyFo~\cite{dyfo2025} uses an external detector, and ZoomEye~\cite{zoomeye2024} performs iterative patch-tree search. Zoom-Refine~\cite{zoomrefine2025} directly uses the frozen model's own localization prediction, but lets a second answering pass determine the result; an incorrect crop or missing context can therefore invalidate an initially correct answer. Confidence-gated methods~\cite{lookwise2026,lazymcot2026} also inspect only uncertain questions, but with a separate grounding module. SEE uses the same model to sample multiple crop proposals and evaluates their answers alongside the original prediction. As summarized in Table~\ref{tab:position}, it combines self-generated evidence with answer aggregation, without additional training, attention access, external detectors, or iterative region search.

\paragraph{Textual evidence and answer aggregation.}
TextCoT~\cite{textcot2024} supplies global descriptions to localized inspection, and DC$^2$~\cite{hrbench2024} combines patch descriptions. Both use generated text to support a final answering pass, rather than weighing a text-assisted answer against independent visual predictions. Self-consistency~\cite{selfconsistency2022} and confidence-based variants~\cite{cisc2025,deepconf2025,selfcertainty2025} aggregate alternative reasoning paths, but resampling reasoning alone does not expose finer visual detail than the supplied input view. RTWI~\cite{rtwi2026} addresses unreliable visual evidence by filtering and weighting interleaved image--text reasoning traces; its reliability estimation operates on both cue mining and subsequent reasoning. SEE instead aggregates separately answered evidence views. Its full-image transcription supplements local crops when relevant text or context lies outside them, while retaining the image alongside the transcript. The resulting answer is combined with crop-based and direct predictions using answer-token confidence, without requiring stage-wise trace filtering. A confidence gate limits additional inspection to uncertain direct answers.

\section{Method}
\label{sec:method}

\begin{figure*}[t]
  \centering
  \includegraphics[width=0.99\textwidth]{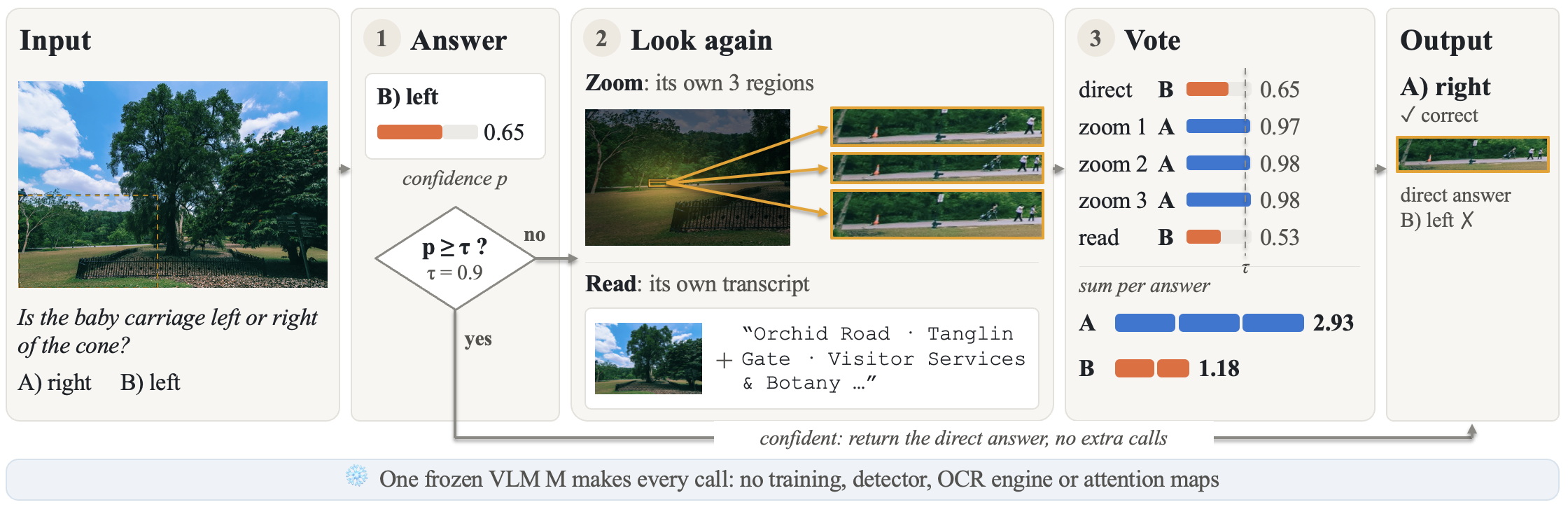}
  \caption{\textbf{Overview of SEE.} \emph{Step 1:} the frozen model $M$ answers the whole-image view $v_I$; if its
  confidence $p_{v_I}$ reaches $\tau$, SEE returns this direct answer. \emph{Step 2:} otherwise SEE elicits evidence with
  two operations: \texttt{Zoom} enlarges the $K$ regions $r_k$ that the model proposes when asked where to look, giving
  the zoomed views $v_k$, and \texttt{Read} has the model transcribe the image, giving the reading view $v_R$.
  \emph{Step 3:} the model answers every view, and each view votes for its answer $y_v$ with its confidence $p_v$
  (Eq.~\ref{eq:see}). One frozen model, no training.}
  \label{fig:overview}
\end{figure*}

As illustrated in Figure~\ref{fig:overview}, given an image $I$ and a question $Q$, SEE uses a frozen VLM $M$ to answer in three steps:
(1) answer the original image and stop if its confidence (Eq.~\ref{eq:conf}) reaches a threshold;
(2) otherwise, elicit additional evidence through \texttt{Zoom} and \texttt{Read};
and (3) aggregate the answers across views.
Algorithm~\ref{alg:see} specifies the complete procedure.

\subsection{Direct Answer and Confidence Gating}
\label{sec:gate}

In Step 1, SEE answers the original whole-image view $v_I=(I,Q)$.
A \emph{view} $v$ consists of an image and a text prompt: the question, preceded in the reading view by the transcript.
We write $M(v)$ for answering a view, which produces an answer $y_v$ and
a confidence $p_v\in(0,1]$. Confidence is defined as the geometric mean
of the probabilities of the $T$ answer tokens:
\begin{equation}
p_v = \exp\Big(\frac{1}{T}\sum_{t=1}^{T}\log p\big(y_{v,t} \mid y_{v,<t},\, v\big)\Big).
\label{eq:conf}
\end{equation}
If the initial confidence satisfies $p_{v_I}\ge\tau$, SEE returns $y_{v_I}$
without further model calls. Otherwise, it retains the initial answer and
its confidence and proceeds to evidence elicitation.

The same confidence definition applies to every answering call.
For multiple-choice questions, the answer is a single option
letter ($T{=}1$), so $p_v$ is the probability assigned to that letter.
For open-ended questions, the model generates a short answer span, and
Eq.~\ref{eq:conf} is evaluated over its tokens.
In \emph{plain mode}, the model answers without a reasoning trace.
In \emph{thinking mode}, it first produces its native reasoning trace; only the final answer
tokens contribute to the confidence average.

\subsection{Evidence Elicitation}
\label{sec:see}\label{sec:evidence}\label{sec:phi}

In Step 2, SEE uses the same frozen model to construct additional views
through two operations: \texttt{Zoom} exposes selected regions at a larger
scale, while \texttt{Read} makes image text available as language tokens.

\paragraph{Region proposals (\texttt{Zoom}).}
Useful localization can remain available even when the model's direct answer
is wrong. SEE elicits this information by asking $M$ where to look, sampling
$K$ responses (temperature $0.8$) to a prompt requesting box coordinates:
\begin{equation}
r_k \sim M(\texttt{where}\mid I, Q), \qquad k=1,\dots,K,
\label{eq:where}
\end{equation}
where $r_k$ is the smallest box enclosing all boxes in the $k$-th response;
responses without a box are discarded. Sampling allows SEE to consider
multiple proposals rather than commit to a single localization prediction.
The localization prompt does not include the initial answer, so a wrong first answer cannot bias where the model looks.
Each valid proposal defines a zoomed view:
\begin{equation}
v_k = \big(\texttt{Zoom}(I, r_k),\; Q\big),
\label{eq:phi}
\end{equation}
where $\texttt{Zoom}$ expands the box by $30\%$ on each side, crops the original
image at full resolution, and upsamples the crop to a shorter side of $448$ pixels if it is smaller.

\paragraph{Transcription-conditioned view (\texttt{Read}).}
A spatial crop can omit text or layout needed to answer a document question.
\texttt{Read} instead asks $M$ to transcribe the full image and places the
resulting text $t$ before the question:
\begin{equation}
\begin{aligned}
t &= \texttt{Read}(I) = M(\texttt{transcribe}\mid I), \\
v_R &= \big(I,\; t \oplus Q\big).
\end{aligned}
\label{eq:read}
\end{equation}
Here $\oplus$ denotes text concatenation. The view retains the original image and its layout while making the
transcribed content available as language tokens. Localization and transcription use no thinking in either answering mode. Prompts and decoding settings are provided in App.~A.

\subsection{Answer Aggregation}
\label{sec:arb}

In Step 3, the model answers each additional view separately and computes
its confidence using Eq.~\ref{eq:conf}. Since a crop may omit relevant content
and a transcript may contain errors, SEE aggregates these predictions with
the original answer rather than letting one additional view decide.
The initial answer $y_{v_I}$ and its confidence $p_{v_I}$ are reused,
without another answering call on the original view.

The views form
$\mathcal{V} = \{v_I\} \cup \{v_k\}_{k} \cup \{v_R\}$,
comprising $2$ to $K{+}2$ views (direct and reading views, plus one per valid proposal).
Each view votes for its answer with its confidence:
\begin{equation}
A^{\star} = \arg\max_{y}\; \sum_{v\in\mathcal{V}:\; y_v = y} p_v .
\label{eq:see}
\end{equation}
Open-ended answers are compared after lower-casing and removing punctuation. Ties are resolved by the earliest view in the order
$v_I, v_1,\dots,v_K, v_R$.
The original prediction retains its vote and is replaced only when another
answer receives greater total support.

\begin{algorithm}[t]
\caption{SEE with a frozen VLM $M$}
\label{alg:see}
\small
\algrenewcommand\algorithmiccomment[1]{\hfill{\color{green!50!black}\textit{// #1}}}
\begin{algorithmic}[1]
\Require image $I$, question $Q$; number of regions $K$; threshold $\tau$
\State $v_I\gets(I,Q)$;\ \ $y_{v_I},\,p_{v_I}\gets M(v_I)$ \Comment{Step 1: answer}
\If{$p_{v_I}\ge\tau$} \Return $y_{v_I}$ \Comment{confident: stop}
\EndIf
\State $\mathcal{V}\gets\{v_I\}$
\For{$k=1,\dots,K$} \Comment{Step 2: \texttt{Zoom}}
  \State $r_k\sim M(\texttt{where}\mid I,Q)$ \Comment{Eq.~\ref{eq:where}}
  \If{$r_k$ has a box} $\mathcal{V}\gets\mathcal{V}\cup\{v_k\}$ \Comment{$v_k$: Eq.~\ref{eq:phi}}
  \EndIf
\EndFor
\State $t\gets M(\texttt{transcribe}\mid I)$ \Comment{Step 2: \texttt{Read}}
\State $\mathcal{V}\gets\mathcal{V}\cup\{(I,\,t\oplus Q)\}$ \Comment{$v_R$, Eq.~\ref{eq:read}}
\For{$v\in\mathcal{V}\setminus\{v_I\}$} \Comment{Step 3: answer the new views}
  \State $y_v,\,p_v\gets M(v)$ \Comment{Eq.~\ref{eq:conf}}
\EndFor
\State \Return $\arg\max_y \sum_{v:\,y_v=y} p_v$ \Comment{vote, Eq.~\ref{eq:see}}
\end{algorithmic}
\end{algorithm}

\section{Experiments}
\label{sec:experiments}

We evaluate SEE across five frozen backbones on high-resolution visual question answering and document understanding.
The experiments answer four questions, in the order the method raises them: does the model know where to look
(Sec.~\ref{sec:elicit}), does SEE beat other training-free methods (Sec.~\ref{sec:main_results}), why does it work
(Sec.~\ref{sec:why}), and finally, does it still help strong models, and at what cost (Sec.~\ref{sec:thinking})? 

\begin{figure*}[!t]
\centering
\setlength{\tabcolsep}{1pt}
\begin{tabular}{ccc}
\includegraphics[width=0.32\textwidth]{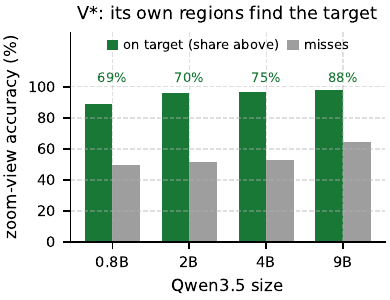} &
\includegraphics[width=0.32\textwidth]{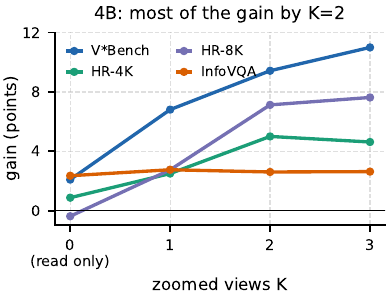} &
\includegraphics[width=0.32\textwidth]{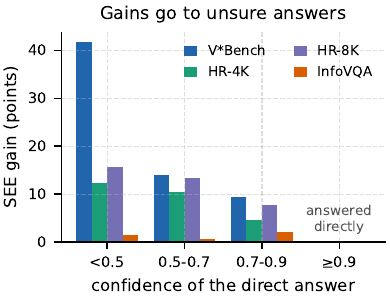} \\
\end{tabular}
\caption{\textbf{Why looking again works} (Qwen3.5 0.8B--9B, no thinking, full sets; 27B: App.~D). \emph{Left:} on V*Bench, zooms whose region covers
the target (green) are far more accurate than those that miss it (grey), and most regions do cover it (share on top).
\emph{Middle:} SEE's gain against the number of zooms $K$ (4B; $K{=}0$: reading view only); most of it arrives with
two. \emph{Right:} SEE's gain by the confidence of the direct answer: largest when the model is unsure, and zero above
the gate.}
\label{fig:strengths}
\end{figure*}

\begin{table*}[t]
  \centering
  \caption{\textbf{Training-free methods on five frozen backbones} (higher is better; no thinking). SEE: when unsure,
the frozen model zooms into regions it proposes and reads its own transcript, then each view votes with its confidence.
Each cell averages one backbone's accuracy on V*Bench and HR-Bench 4K/8K and ANLS on InfographicVQA; Q3.5: Qwen3.5.
SEE has the best small-object average (Small$^{\circ}$, $15$ cells), significantly above every baseline's (paired
bootstrap, App.~B), ties self-consistency overall (Overall$^{\circ}$, $20$ cells), and lifts 4B ($84.6$) above the 9B
direct answer ($80.5$). Shading ranks each column (darker is better); bold\,/\,underline: best\,/\,second; per benchmark:
App.~D.}
  \label{tab:main-avg}
  \begingroup
  \fontsize{8}{9.6}\selectfont
  \setlength{\tabcolsep}{3pt}
  \renewcommand{\arraystretch}{1.16}
  \begin{tabular*}{\textwidth}{
    @{\extracolsep{\fill}} l *{5}{r}
    >{\columncolor{cvprOverallGray}}r >{\columncolor{cvprOverallGray}}r @{}
  }
    \toprule
    Method & Q3.5-0.8B~\cite{qwen2025} & Q3.5-2B~\cite{qwen2025} & Q3.5-4B~\cite{qwen2025} & Q3.5-9B~\cite{qwen2025} & GLM-4.1V~\cite{glm41v} & Overall$^{\circ}$ & Small$^{\circ}$ \\
    \midrule
    Direct & \cvprTableHeat{7}{62.8} & \cvprTableHeat{10}{72.8} & \cvprTableHeat{10}{77.9} & \cvprTableHeat{8}{80.5} & \cvprTableHeat{3}{73.2} & \cvprTableHeat{7}{73.4} & \cvprTableHeat{7}{72.4} \\
    Self-consistency~\cite{selfconsistency2022} & \cvprTableHeat{17}{\cvprTableSecond{68.7}} & \cvprTableHeat{21}{\cvprTableBest{75.8}} & \cvprTableHeat{17}{\cvprTableSecond{83.5}} & \cvprTableHeat{16}{\cvprTableSecond{84.3}} & \cvprTableHeat{7}{75.3} & \cvprTableHeat{17}{\cvprTableSecond{77.5}} & \cvprTableHeat{14}{75.9} \\
    Zoom-Refine~\cite{zoomrefine2025} & \cvprTableHeat{10}{64.7} & \cvprTableHeat{3}{66.1} & \cvprTableHeat{14}{80.4} & \cvprTableHeat{12}{82.9} & \cvprTableHeat{17}{\cvprTableSecond{76.6}} & \cvprTableHeat{10}{74.1} & \cvprTableHeat{10}{74.7} \\
    ZoomEye~\cite{zoomeye2024} & \textit{60.8}  & \textit{70.3}  & \textit{78.4}  & \textit{84.2}  & \textit{72.5}  & \textit{73.2}  & \textit{72.8}  \\
    HiDe~\cite{hide2025} & \cvprTableHeat{14}{66.7} & \cvprTableHeat{14}{73.1} & \cvprTableHeat{7}{76.5} & \cvprTableHeat{21}{\cvprTableBest{88.2}} & \cvprTableHeat{21}{\cvprTableBest{77.9}} & \cvprTableHeat{14}{76.5} & \cvprTableHeat{17}{\cvprTableSecond{76.4}} \\
    DyFo~\cite{dyfo2025} & \cvprTableHeat{3}{62.6} & \cvprTableHeat{7}{70.3} & \cvprTableHeat{3}{70.9} & \cvprTableHeat{3}{79.0} & \cvprTableHeat{10}{75.8} & \cvprTableHeat{3}{71.7} & \cvprTableHeat{3}{71.4} \\
    \midrule
    \rowcolor{cvprOursGray}
    \textbf{SEE (Ours)} & \cvprTableHeat{21}{\cvprTableBest{69.0}} & \cvprTableHeat{17}{\cvprTableSecond{75.7}} & \cvprTableHeat{21}{\cvprTableBest{84.6}} & \cvprTableHeat{16}{\cvprTableSecond{84.3}} & \cvprTableHeat{14}{76.1} & \cvprTableHeat{21}{\cvprTableBest{78.0}} & \cvprTableHeat{21}{\cvprTableBest{78.2}} \\
    \bottomrule
  \end{tabular*}
  \endgroup
\end{table*}

\begin{table}[t]
\centering
\caption{\textbf{How often each method fixes or breaks the direct answer} (\%, small-object benchmarks pooled). Each cell reads
\emph{recovery\,/\,damage}: $22.0\,/\,2.5$ means a method fixes $22.0\%$ of the direct answer's errors and breaks
$2.5\%$ of its right answers; the ideal is high recovery at zero damage (the direct answer itself is $0\,/\,0$). No method breaks fewer right answers than SEE on
any backbone (HiDe ties it on GLM-4.1V). Bold: lowest damage per backbone. No thinking, full sets; SEE per benchmark: App.~D.}
\setlength{\tabcolsep}{3pt}
\resizebox{\columnwidth}{!}{%
\begin{tabular}{lccccc}
\toprule
\textbf{Method (recovery / damage)} & \textbf{0.8B} & \textbf{2B} & \textbf{4B} & \textbf{9B} & \textbf{GLM-4.1V} \\
\midrule
Self-consistency & $26.0\,/\,9.3$ & $21.3\,/\,5.8$ & $43.4\,/\,5.2$ & $33.0\,/\,4.8$ & $18.9\,/\,5.7$ \\
Zoom-Refine      & $28.3\,/\,9.4$ & $25.4\,/\,16.4$ & $33.9\,/\,5.9$ & $33.7\,/\,4.4$ & $35.9\,/\,7.3$ \\
HiDe             & $41.2\,/\,15.2$ & $45.7\,/\,17.0$ & $53.3\,/\,18.9$ & $58.1\,/\,3.7$ & $31.2\,/\,\mathbf{4.7}$ \\
DyFo             & $24.5\,/\,22.5$ & $39.2\,/\,21.7$ & $36.7\,/\,26.1$ & $37.2\,/\,11.1$ & $30.0\,/\,7.1$ \\
\midrule
\rowcolor{gray!15}
\textbf{SEE}     & $22.0\,/\,\mathbf{2.5}$ & $19.3\,/\,\mathbf{2.2}$ & $36.7\,/\,\mathbf{2.4}$ & $27.2\,/\,\mathbf{2.1}$ & $21.3\,/\,\mathbf{4.7}$ \\
\bottomrule
\end{tabular}}
\label{tab:recdmg}
\end{table}

\subsection{Experimental Setup}
\paragraph{Models and benchmarks.} SEE needs a VLM that can write boxes as text, read text in an image and return token
probabilities. We use Qwen3.5~\cite{qwen2025} at 0.8B, 2B, 4B and 9B, one recipe at four sizes, to see whether the gap
closes with scale, and GLM-4.1V-9B-Thinking~\cite{glm41v} (GLM-4.1V), a reasoning model from another family, to see
whether SEE transfers; all are frozen. The benchmarks cover the two regimes SEE targets. \emph{Small objects in
high-resolution images:} V*Bench~\cite{vstar2024} ($191$ multiple-choice questions) and HR-Bench 4K and
8K~\cite{hrbench2024} ($800$ each). \emph{Dense text and layout:} InfographicVQA~\cite{infographicvqa2021} ($500$
open-ended validation questions, a fixed random sample), scored with ANLS, which gives partial credit to near-misses.

\paragraph{Baselines.} We compare five training-free methods as listed in Table~\ref{tab:position}. Our self-consistency baseline adapts~\cite{selfconsistency2022}: it samples three chains that describe where to look, what to zoom into, and what to read, then applies SEE's gate and confidence-weighted vote. Zoom-Refine~\cite{zoomrefine2025} and ZoomEye~\cite{zoomeye2024} follow their published recipes; HiDe~\cite{hide2025} uses its released code, and DyFo~\cite{dyfo2025} uses GroundingDINO. We do not tune baseline settings on our data. A baseline that runs out of GPU memory would fall back to the direct answer as final output. Implementation details are listed in App.~E.

\begin{figure}[t]
\centering
\includegraphics[width=0.95\columnwidth]{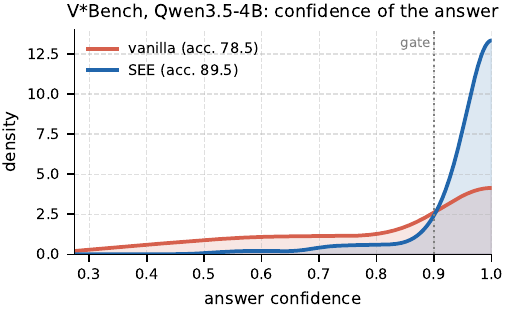}
\caption{\textbf{Answer confidence before and after SEE} (V*Bench, Qwen3.5-4B; a second 4B run, App.~B). For SEE, the
confidence is the highest among the views that vote for its answer; answers above the gate ($\tau{=}0.9$) are returned
unchanged. $47\%$ of the direct answers fall below $0.8$, where most errors are, against $10\%$ after SEE's vote;
accuracy rises from $78.5\%$ to $89.5\%$.}
\label{fig:conf}
\end{figure}

\paragraph{SEE Protocol.} \label{sec:setup}\label{sec:impl}\label{sec:thinkmode}
One setting serves every model and benchmark: $K{=}3$ regions and the gate $\tau{=}0.9$. $K$, the margin and the crop
size were chosen once, on $60$ items per benchmark at 4B; removing these items moves every gain by at most $1.2$ points. The vote rule and $\tau$ are not held out but compared on the full sets. Gains over the direct answer are tested with exact McNemar or paired sign-flip tests, and differences between
methods with a paired bootstrap over questions. Cost is wall-clock time per question, as a multiple of the direct answer
in the same run. At 4B, SEE ran four times: Tables~\ref{tab:main-avg}--\ref{tab:controls} use one run (the
\emph{comparison run}), Table~\ref{tab:ablation} and Figs.~\ref{fig:teaser},~\ref{fig:strengths} and~\ref{fig:conf}
another (the \emph{analysis run}). Decoding settings and all four runs are in App.~A and~B.

\begin{table}[t]
\centering
\caption{\textbf{The gain comes from where the model looks} (Qwen3.5-4B, no thinking). The controls keep all of
SEE (gate, vote, reading view) but replace the model's own regions with random crops of the same sizes, or with a
fixed $2{\times}2$ grid. Random crops fall below the direct answer on every benchmark ($-2.9$ on average), and the
grid recovers only $1.5$ of SEE's $6.7$ points. Self-consistency and HiDe for reference; \emph{Text}: uses
the reading view. Calls/question: direct $1$, self-consistency $7$, HiDe $3$, SEE $1$ or $9$. Bold: best
per column.}
\setlength{\tabcolsep}{2.5pt}
\resizebox{\columnwidth}{!}{%
\begin{tabular}{lcccccc}
\toprule
\textbf{Method} & \textbf{Text} & \textbf{V*} & \textbf{HR-4K} & \textbf{HR-8K} & \textbf{InfoVQA} & \textbf{Mean} \\
\midrule
vanilla                                 & & 78.5 & 77.6 & 73.4 & 82.0 & 77.9 \\
self-consistency~\cite{selfconsistency2022} & & 82.2 & \textbf{84.9} & 80.0 & \textbf{87.0} & 83.5 \\
HiDe~\cite{hide2025} & & 73.3 & 76.0 & 73.1 & 83.7 & 76.5 \\
\rowcolor{gray!15}
\textbf{SEE}                            & \checkmark & \textbf{90.1} & 82.6 & \textbf{81.4} & 84.2 & \textbf{84.6} \\
\midrule
\multicolumn{7}{l}{\emph{controls: SEE with other zoomed views}} \\
\quad random crops                      & \checkmark & 73.8 & 74.6 & 70.0 & 81.7 & 75.0 \\
\quad $2{\times}2$ tiles                & \checkmark & 81.7 & 78.0 & 73.9 & 83.9 & 79.4 \\
\bottomrule
\end{tabular}}
\label{tab:controls}
\end{table}

\begin{table}[t]
\centering
\caption{\textbf{What each part of SEE contributes} (Qwen3.5-4B, no thinking; a second 4B run, App.~B). Rows add SEE's parts to the direct
answer. \texttt{Zoom} helps small objects but hurts InfographicVQA, and \texttt{Read} does the reverse; voting over
both keeps both gains, and the gate halves the time. Time: mean seconds per question, counting a gated stage only when
it runs. Bold: best per column.}
\setlength{\tabcolsep}{3pt}
\resizebox{\columnwidth}{!}{%
\begin{tabular}{lcccccc}
\toprule
\textbf{Candidates / rule} & \textbf{V*} & \textbf{HR-4K} & \textbf{HR-8K} & \textbf{InfoVQA} & \textbf{Mean} & \textbf{Time} \\
\midrule
direct answer only               & 78.5 & 77.6 & 73.4 & 82.0 & 77.9 & 7.4 \\
+ \texttt{Zoom}                  & \textbf{90.1} & 80.4 & 80.8 & 79.8 & 82.8 & 55.2 \\
+ \texttt{Read}                  & 80.6 & 78.5 & 73.0 & 84.2 & 79.1 & 34.6 \\
+ both, vote (Eq.~\ref{eq:see})  & 89.0 & 81.4 & 80.9 & \textbf{85.3} & 84.1 & 82.4 \\
\rowcolor{gray!15}
+ gate (\textbf{SEE})            & 89.5 & \textbf{82.3} & \textbf{81.0} & 84.7 & \textbf{84.4} & 43.5 \\
\bottomrule
\end{tabular}}
\label{tab:ablation}
\end{table}

\subsection{Does the model know where to look?}
\label{sec:elicit}

SEE rests on one premise: when the direct answer is wrong, the model's own regions often still contain the evidence. We test this premise on V*Bench using its target annotations. Among the $41$ questions that Qwen3.5-4B answers incorrectly, the medoid of five localization requests gives a box with IoU ${\geq}0.5$ on $20$. We call this discrepancy the \emph{elicitation gap}: the model can provide a useful location even when its direct answer is wrong.

We then examine the regions sampled by SEE across model sizes.
Here, we measure coverage as the fraction of regions containing at least half
of the annotated target, rather than IoU. Coverage increases from $69\%$ at 0.8B to $88\%$ at 9B, and zoomed views that cover the target yield substantially higher answer accuracy than those that miss it, as shown in the left part of Figure~\ref{fig:strengths}. Larger models also benefit from inspecting these regions: SEE improves the 9B model by $3.6$--$6.2$ points across the small-object benchmarks, and Qwen3.5-27B by $1.8$--$3.2$ (App.~D). The results indicate that localization runs
ahead of recognition. It can often point at evidence it cannot yet resolve.

\begin{figure}[t]
  \centering
  \includegraphics[width=\columnwidth]{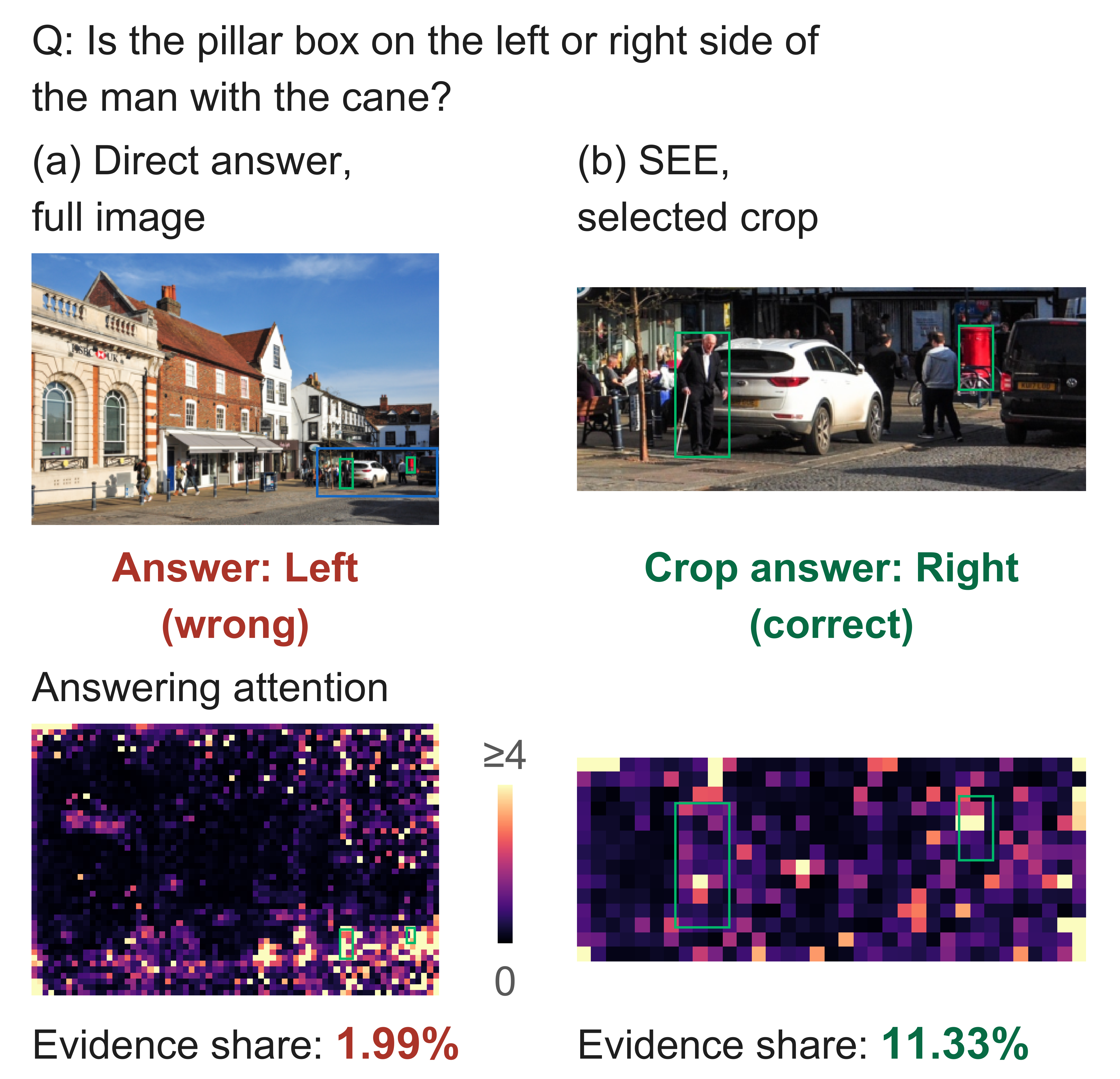}
    \caption{\textbf{Answering attention before and after zooming.}
    Qwen3.5-4B answers incorrectly from the full image and correctly
    from SEE's crop. The referenced objects receive $1.99\%$ and
    $11.33\%$ of each view's image attention (the evidence share), respectively.
    Blue marks the crop; green marks the referenced objects.}
  \label{fig:detail-inspection}
\end{figure}

\begin{figure*}[t]
\centering
\setlength{\tabcolsep}{2pt}
\renewcommand{\arraystretch}{0.6}
\begin{tabular}{cc}
\includegraphics[width=0.44\textwidth]{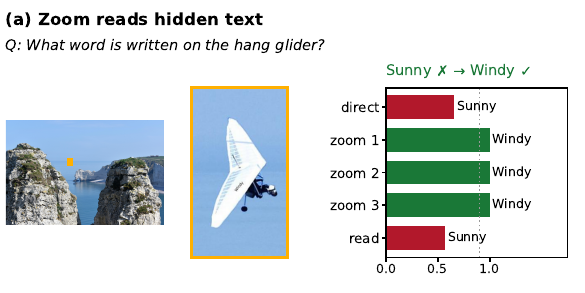} &
\includegraphics[width=0.44\textwidth]{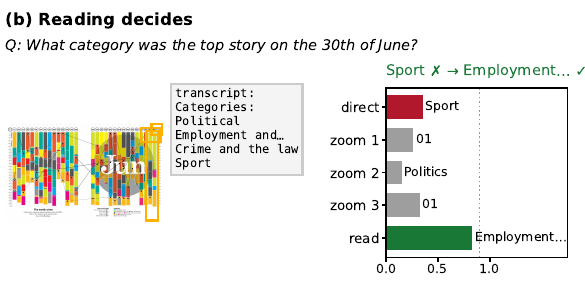} \\
\includegraphics[width=0.44\textwidth]{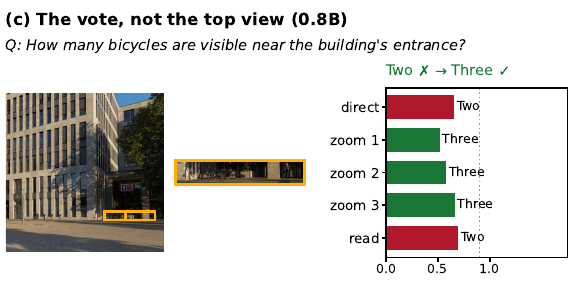} &
\includegraphics[width=0.44\textwidth]{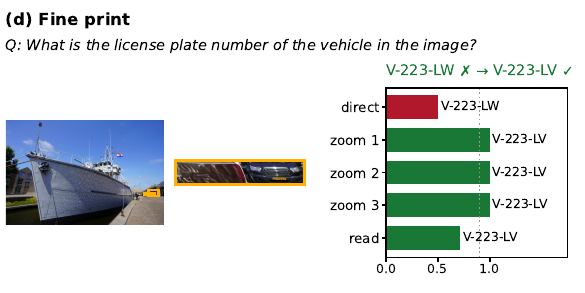} \\
\includegraphics[width=0.44\textwidth]{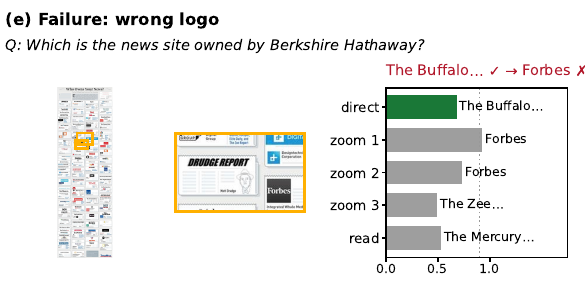} &
\includegraphics[width=0.44\textwidth]{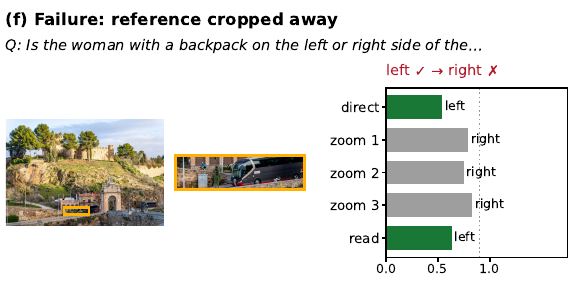} \\
\end{tabular}
\caption{\textbf{How SEE decides, on real questions} (Qwen3.5-4B unless marked 0.8B). Each cell shows the image with the
model's own regions (yellow), the evidence that decided (the winning zoom or the transcript), and every view's confidence (green: right; red: the direct
answer's wrong answer; grey: another wrong answer; dotted: the gate $\tau$); above the bars, \emph{direct answer}
$\rightarrow$ \emph{SEE's answer}. The model's regions find the target (a, d), the transcript settles text (b), and
agreeing zooms outvote the most confident view (c). Failures: two confident zooms on the wrong logo (e), or a crop
that keeps only one of the two objects being compared (f). Cases were mined automatically by category; more in
App.~C.}
\label{fig:examples}
\end{figure*}

\subsection{Does SEE beat other training-free methods?}
\label{sec:main_results}

Table~\ref{tab:main-avg} compares SEE with the five baselines on every backbone. We have some key observations:

\paragraph{SEE gains the most on small objects.} 
SEE improves every small-object benchmark on every backbone by
$2.3$--$13.1$ points, the only compared method to do so. 
SEE achieves the highest small-object average of $78.2$, significantly
outperforming every baseline. These gains persist across runs and after excluding the tuning items. The improvements extend to the larger backbones, suggesting that additional visual evidence remains useful as model capacity increases.

\paragraph{SEE rarely breaks a right answer.}
Table~\ref{tab:recdmg} decomposes answer revisions into \emph{recovery},
the fraction of incorrect direct answers corrected, and \emph{damage}, the fraction of correct direct answers overturned. SEE achieves the lowest damage rate on every backbone, ranging from $2.1\%$ to $4.7\%$, with HiDe matching it on GLM-4.1V. Higher recovery does not necessarily yield better final accuracy:
at 4B, HiDe corrects $53.3\%$ of the initial errors but overturns $18.9\%$ of correct answers, compared with SEE's $36.7\%$ recovery and $2.4\%$ damage. Zoom-Refine, ZoomEye and HiDe let a single final pass decide, and DyFo weights its votes by crop size rather than by the model's confidence. SEE keeps the direct answer as one vote and replaces it only when the new views outweigh it.

\paragraph{SEE enables smaller models to rival larger ones.}
On small-object benchmarks, SEE enables smaller Qwen3.5 models to approach or exceed the direct-answer performance of larger ones. Table~\ref{tab:modes} shows that 0.8B with SEE nearly matches 2B, 2B surpasses 4B, and 4B exceeds 9B by $6.1$ points. SEE makes evidence acquisition explicit: the model identifies relevant regions and inspects the resulting views before combining their answers. This structure provides smaller models with a guided visual reasoning
process, while allowing larger models to revisit evidence missed during
direct answering.
The gains at both ends of the model range suggest that explicit evidence
acquisition can partly compensate for limited model capacity and remains
useful as capacity increases.
This comes at an inference-time cost: 4B with SEE uses fewer than half
the parameters of 9B but takes approximately $4.3\times$ as long.

\subsection{Why does SEE work?}
\label{sec:why}\label{sec:controls}\label{sec:compare}\label{sec:text}\label{sec:arb_ablation}\label{sec:ablation}\label{sec:unified}\label{sec:read}

\paragraph{It is where the model looks.} If the gain came from extra views or from enlarging the image, any crops would
do. Table~\ref{tab:controls} keeps all of SEE but replaces the model's regions with random crops of the same sizes, or
with a fixed $2{\times}2$ grid. Random crops make SEE worse than the direct answer, and the grid recovers only $1.5$ of
SEE's $6.7$ points; the rest comes from the regions the model chooses.

\paragraph{SEE makes visual evidence more prominent.}
Figure~\ref{fig:detail-inspection} compares answering-stage attention
on the full image and SEE's selected crop. The two referenced objects receive only $1.99\%$ of image attention in the full-image view, compared with $11.33\%$ in the crop. Enlarging the evidence gives it more visual tokens and a larger share of the attention allocated to the image, alongside a correct crop
answer where direct answering fails. This pattern is consistent with improved access to fine-grained visual cues, although cropping also changes resolution and context.

\paragraph{SEE alleviates attention dilution.}
Small objects occupy few visual tokens, so even preferential attention to those tokens can leave them with little weight in the full-image view.
SEE enlarges these objects and removes much of the surrounding scene, increasing their visual-token representation.
This increase occurs despite lower relative attention density, suggesting that expanding the evidence's representation can alleviate its dilution within the full image. The example helps explain why localization alone may be insufficient: decisive details can remain underrepresented during answering even when the model can locate them.

\paragraph{Each operation serves its own regime.} Table~\ref{tab:ablation} adds SEE's parts one at a time at 4B.
\texttt{Zoom} alone lifts V*Bench by $+11.6$ points but lowers InfographicVQA, where crops cut the page apart and zoomed
views are overconfident; \texttt{Read} alone does the reverse. Other zoom methods have no such safeguard and lose up to
$12.7$ ANLS on InfographicVQA, while SEE never drops more than $1.3$ below the direct answer (App.~D). What helps is
the content of the transcript, not a longer prompt: the model's own transcript adds $+2.2$ ANLS, an empty one $-0.2$ and
another image's $-1.1$ (App.~B).

\paragraph{The vote keeps both gains.} Voting over the direct answer, the zooms and the reading view keeps the gains of
both operations (Table~\ref{tab:ablation}). No single view does: keeping only the most confident one is stronger on
small objects but weaker on documents at every size (App.~B). Most of the gain arrives with two zooms
(Fig.~\ref{fig:strengths}, middle).

\paragraph{Confidence decides when to look and which view to trust.} SEE's gain sits where the direct answer is unsure
and vanishes above the gate (Fig.~\ref{fig:strengths}, right), which is why the gate can skip confident questions and
halve the time at no loss (Table~\ref{tab:ablation}). When SEE does look again, it replaces unsure direct answers with
views the model answers decisively (Fig.~\ref{fig:conf}), and agreement is the signal it relies on: when all three zooms
agree, it is right $87\%$ of the time on V*Bench, against $46$--$48\%$ when at most one does.

\begin{table}[t]
\centering
\caption{\textbf{SEE on top of native thinking, on every backbone} (higher is better). \emph{plain}: the direct answer
of Table~\ref{tab:main-avg}; \emph{thinking}: the model reasons first; each without and with SEE, plus self-consistency
(three extra thinking chains). On small objects, SEE adds $+1.9$ to $+2.6$ points on top of thinking, more than
self-consistency; on InfographicVQA, self-consistency is stronger. Bold: best per column and block.}
\setlength{\tabcolsep}{4pt}
\resizebox{\columnwidth}{!}{%
\begin{tabular}{lccccc}
\toprule
\textbf{Answering mode} & \textbf{Q3.5-0.8B} & \textbf{2B} & \textbf{4B} & \textbf{9B} & \textbf{GLM-4.1V} \\
\midrule
\multicolumn{6}{l}{\emph{Small objects: mean accuracy over V*Bench, HR-Bench 4K and 8K}} \\
plain & 64.5 & 72.7 & 76.5 & 78.6 & 69.7 \\
plain + \textbf{SEE} & 72.5 & 77.1 & 84.7 & 83.3 & \textbf{73.6} \\
thinking & 71.1 & 77.9 & 84.9 & 86.4 & 72.2 \\
thinking + self-consistency & 73.0 & 79.4 & 86.0 & 87.0 & 72.2 \\
thinking + \textbf{SEE} & \textbf{73.3} & \textbf{80.5} & \textbf{87.3} & \textbf{88.4} & 72.1 \\
\midrule
\multicolumn{6}{l}{\emph{InfographicVQA (ANLS)}} \\
plain & 57.6 & 72.8 & 82.0 & 85.9 & 83.7 \\
plain + \textbf{SEE} & 58.6 & 71.5 & 84.2 & 87.2 & 83.8 \\
thinking & 71.3 & 82.5 & 92.4 & 93.0 & 87.4 \\
thinking + self-consistency & \textbf{74.2} & \textbf{85.1} & \textbf{93.0} & \textbf{93.2} & \textbf{87.7} \\
thinking + \textbf{SEE} & 71.9 & 82.5 & 92.9 & 92.7 & 86.9 \\
\bottomrule
\end{tabular}}
\label{tab:modes}
\end{table}

\paragraph{What it looks like.} Fig.~\ref{fig:examples} shows SEE on real questions. The model's own regions find a small
target that the direct answer misses (a, d), its transcript settles a question about text (b), and three agreeing zooms
outvote a single, more confident view (c). The same mechanisms explain the failures: two confident zooms on the wrong logo
(e), or a crop that keeps only one of the two objects a relation question compares (f), can overturn a right answer.
\emph{Takeaway:} each operation covers one failure of a single look, and the model's own confidence decides when to look
again and which view to trust.

\subsection{Does SEE help strong models, and at what cost?}
\label{sec:thinking}\label{sec:cost}

\paragraph{On top of native thinking.} Thinking gathers evidence in text, over the pixels the model already has; SEE
gives it new pixels and tokens. Table~\ref{tab:modes} runs both on every backbone. Thinking lifts every Qwen3.5 model on
small objects, and SEE adds a further $+1.9$ to $+2.6$ points on top, more than three extra thinking chains
(self-consistency); with both, 4B edges out 9B with thinking. On InfographicVQA, reasoning over the page remains stronger, and with thinking GLM-4.1V is confident
on almost every question, so SEE's gate rarely opens. The two add up because they act on different things: thinking
helps the model reason about what it sees, while SEE changes what it sees.

\paragraph{Cost.} SEE makes its extra calls in parallel and only for unsure questions. At 4B it takes $5.9\times$ the time
of a direct answer, close to self-consistency ($5.0\times$) and far below DyFo ($20\times$) and ZoomEye's tree search
($127\times$; Table~\ref{tab:position}); across backbones it takes $3.4$--$6.7\times$, and on top of thinking $1.6\times$
thinking alone. \emph{Takeaway:}
looking again adds to reasoning, at a cost close to that of self-consistency.

\section{Conclusion}
Frozen VLMs often know where the answer is even when they get it wrong. SEE puts this to use with two basic operations,
zooming where the model points and reading what the image says, and a confidence-weighted vote that keeps the direct
answer in the decision, with no training and no external tools. Three lessons stand out. First, the model's own regions
are the evidence: random crops in their place turn SEE's gain into a loss. Second, keeping the direct answer in the vote is
what makes looking again safe: SEE improves every small-object benchmark on every backbone while rarely breaking a right
answer. Third, looking again and reasoning are complementary: SEE adds to native thinking, while reasoning over the whole
page remains stronger on dense text. SEE is a natural choice when the details are small, the model is frozen and its token
probabilities are available: on small objects it is worth about one model size (9B with SEE matches the 27B direct answer; App.~D). Attention-based zooming is cheaper and stronger at 9B; combining it with SEE without giving up black-box access is next.

\paragraph{Limitations.} SEE needs a model that can point at what it is asked about: GLM-4.1V's regions cover the
V*Bench target less often than Qwen3.5-9B's ($61\%$ vs.\ $88\%$). Gains concentrate on single objects (larger than on
relations in $9$ of $10$ HR-Bench cells; App.~B), as a zoom can crop away the second object. The gate and the vote are only as good as
the confidence: thinking GLM-4.1V is confident on almost every question, so SEE rarely acts, and a confidently wrong view
can overturn a right answer (Fig.~\ref{fig:examples}e,f). The vote rule and $\tau$ were compared on the full sets; a test set untouched by every choice would settle
how much this matters. Multi-image input, video and long answers are untested.

\section*{LLM Usage Statement}
We used Claude (Anthropic) to assist with the experiments (evaluation, baseline and job scripts; building tables and figures for the experiments) and with formatting and rewording the text. All numbers come from our own runs and all the text was written by the authors having it reviewed and corrected by the LLM; the authors designed the method and experiments and take full responsibility for the content.

{
    \small
    \bibliographystyle{ieeenat_fullname}
    \bibliography{main}

\begin{thebibliography}{28}
\providecommand{\natexlab}[1]{#1}
\providecommand{\url}[1]{\texttt{#1}}
\expandafter\ifx\csname urlstyle\endcsname\relax
  \providecommand{\doi}[1]{doi: #1}\else
  \providecommand{\doi}{doi: \begingroup \urlstyle{rm}\Url}\fi

\bibitem[Fu et~al.(2026)Fu, Wang, Tian, and Zhao]{deepconf2025}
Yichao Fu, Xuewei Wang, Yuandong Tian, and Jiawei Zhao.
\newblock Deep think with confidence.
\newblock In \emph{ICLR}, 2026.

\bibitem[Hong et~al.(2025)]{glm41v}
Wenyi Hong et~al.
\newblock {GLM-4.5V} and {GLM-4.1V-Thinking}: Towards versatile multimodal
  reasoning with scalable reinforcement learning.
\newblock \emph{arXiv preprint arXiv:2507.01006}, 2025.

\bibitem[Kang et~al.(2025{\natexlab{a}})Kang, Kim, Kim, and
  Hwang]{visualsink2025}
Seil Kang, Jinyeong Kim, Junhyeok Kim, and Seong~Jae Hwang.
\newblock See what you are told: Visual attention sink in large multimodal
  models.
\newblock In \emph{ICLR}, 2025{\natexlab{a}}.

\bibitem[Kang et~al.(2025{\natexlab{b}})Kang, Zhao, and
  Song]{selfcertainty2025}
Zhewei Kang, Xuandong Zhao, and Dawn Song.
\newblock Scalable best-of-{N} selection for large language models via
  self-certainty.
\newblock In \emph{NeurIPS}, 2025{\natexlab{b}}.
\newblock arXiv:2502.18581.

\bibitem[Lai et~al.(2026)Lai, Li, Li, Liu, Li, and Zhao]{minio32025}
Xin Lai, Junyi Li, Wei Li, Tao Liu, Tianjian Li, and Hengshuang Zhao.
\newblock {Mini-o3}: Scaling up reasoning patterns and interaction turns for
  visual search.
\newblock In \emph{ICLR}, 2026.

\bibitem[Li et~al.(2025)Li, Xu, Zhao, and Peng]{dyfo2025}
Geng Li, Jinglin Xu, Yunzhen Zhao, and Yuxin Peng.
\newblock {DyFo}: A training-free dynamic focus visual search for enhancing
  {LMMs} in fine-grained visual understanding.
\newblock In \emph{CVPR}, 2025.
\newblock arXiv:2504.14920.

\bibitem[Li et~al.(2026)Li, Yang, Lin, Dai, Yang, and Peng]{rtwi2026}
Haobin Li, Yutong Yang, Yijie Lin, Xiang Dai, Mouxing Yang, and Xi Peng.
\newblock Reliable thinking with images.
\newblock In \emph{ICML}, 2026.
\newblock arXiv:2602.12916.

\bibitem[Liu et~al.(2024{\natexlab{a}})Liu, Zeng, Ren, Li, Zhang, Yang, Jiang,
  Li, Yang, Su, Zhu, and Zhang]{groundingdino2024}
Shilong Liu, Zhaoyang Zeng, Tianhe Ren, Feng Li, Hao Zhang, Jie Yang, Qing
  Jiang, Chunyuan Li, Jianwei Yang, Hang Su, Jun Zhu, and Lei Zhang.
\newblock Grounding {DINO}: Marrying {DINO} with grounded pre-training for
  open-set object detection.
\newblock In \emph{ECCV}, 2024{\natexlab{a}}.

\bibitem[Liu et~al.(2024{\natexlab{b}})Liu, Zheng, and Chen]{pai2024eccv}
Shi Liu, Kecheng Zheng, and Wei Chen.
\newblock Paying more attention to image: A training-free method for
  alleviating hallucination in {LVLMs}.
\newblock In \emph{ECCV}, 2024{\natexlab{b}}.
\newblock arXiv:2407.21771.

\bibitem[Liu et~al.(2025)]{hide2025}
Xianjie Liu et~al.
\newblock {HiDe}: Rethinking the zoom-{IN} method in high resolution {MLLMs}
  via hierarchical decoupling.
\newblock \emph{arXiv preprint arXiv:2510.00054}, 2025.

\bibitem[Liu et~al.(2024{\natexlab{c}})Liu, Yang, Liu, Li, Ma, Zhang, and
  Bai]{textmonkey2024}
Yuliang Liu, Biao Yang, Qiang Liu, Zhang Li, Zhiyin Ma, Shuo Zhang, and Xiang
  Bai.
\newblock {TextMonkey}: An {OCR}-free large multimodal model for understanding
  document.
\newblock \emph{arXiv preprint arXiv:2403.04473}, 2024{\natexlab{c}}.

\bibitem[Luan et~al.(2024)Luan, Feng, Chen, Wang, Zhou, and Li]{textcot2024}
Bozhi Luan, Hao Feng, Hong Chen, Yonghui Wang, Wengang Zhou, and Houqiang Li.
\newblock {TextCoT}: Zoom in for enhanced multimodal text-rich image
  understanding.
\newblock \emph{arXiv preprint arXiv:2404.09797}, 2024.

\bibitem[Mathew et~al.(2022)Mathew, Bagal, Tito, Karatzas, Valveny, and
  Jawahar]{infographicvqa2021}
Minesh Mathew, Viraj Bagal, Rub{\`e}n~P{\'e}rez Tito, Dimosthenis Karatzas,
  Ernest Valveny, and C.~V. Jawahar.
\newblock {InfographicVQA}.
\newblock In \emph{WACV}, 2022.

\bibitem[Peng et~al.(2024)]{kosmos2}
Zhiliang Peng et~al.
\newblock Kosmos-2: Grounding multimodal large language models to the world.
\newblock In \emph{ICLR}, 2024.

\bibitem[{Qwen Team}(2026)]{qwen2025}
{Qwen Team}.
\newblock {Qwen3.5}: Towards native multimodal agents.
\newblock \url{https://qwen.ai/blog?id=qwen3.5}, 2026.

\bibitem[Shen et~al.(2025)Shen, Zhao, Zhao, Xu, Zhang, Zhu, and
  Yin]{zoomeye2024}
Haozhan Shen, Kangjia Zhao, Tiancheng Zhao, Ruochen Xu, Zilun Zhang, Mingwei
  Zhu, and Jianwei Yin.
\newblock {ZoomEye}: Enhancing multimodal {LLMs} with human-like zooming
  capabilities through tree-based image exploration.
\newblock In \emph{EMNLP}, 2025.

\bibitem[Shen et~al.(2026)]{lookwise2026}
Yuxiang Shen et~al.
\newblock {LookWise}: Knowing when and where to look for fine-grained visual
  reasoning in multimodal large language models.
\newblock \emph{arXiv preprint arXiv:2603.00171}, 2026.

\bibitem[Taubenfeld et~al.(2025)Taubenfeld, Sheffer, Ofek, Feder, Goldstein,
  Gekhman, and Yona]{cisc2025}
Amir Taubenfeld, Tom Sheffer, Eran Ofek, Amir Feder, Ariel Goldstein, Zorik
  Gekhman, and Gal Yona.
\newblock Confidence improves self-consistency in {LLMs}.
\newblock In \emph{Findings of ACL}, 2025.

\bibitem[Tong et~al.(2024)Tong, Liu, Zhai, Ma, LeCun, and Xie]{mmvp2024}
Shengbang Tong, Zhuang Liu, Yuexiang Zhai, Yi Ma, Yann LeCun, and Saining Xie.
\newblock Eyes wide shut? exploring the visual shortcomings of multimodal
  {LLMs}.
\newblock In \emph{CVPR}, 2024.

\bibitem[Wang et~al.(2025{\natexlab{a}})Wang, Su, Ren, Lin, and
  Chen]{pixelreasoner2025}
Haozhe Wang, Alex Su, Weiming Ren, Fangzhen Lin, and Wenhu Chen.
\newblock Pixel reasoner: Incentivizing pixel-space reasoning with
  curiosity-driven reinforcement learning.
\newblock In \emph{NeurIPS}, 2025{\natexlab{a}}.

\bibitem[Wang et~al.(2025{\natexlab{b}})]{hrbench2024}
Wenbin Wang et~al.
\newblock Divide, conquer and combine: A training-free framework for
  high-resolution image perception in multimodal large language models.
\newblock In \emph{AAAI}, 2025{\natexlab{b}}.

\bibitem[Wang et~al.(2023)Wang, Wei, Schuurmans, Le, Chi, Narang, Chowdhery,
  and Zhou]{selfconsistency2022}
Xuezhi Wang, Jason Wei, Dale Schuurmans, Quoc Le, Ed Chi, Sharan Narang,
  Aakanksha Chowdhery, and Denny Zhou.
\newblock Self-consistency improves chain of thought reasoning in language
  models.
\newblock In \emph{ICLR}, 2023.

\bibitem[Wang et~al.(2026)]{lazymcot2026}
Yifan Wang et~al.
\newblock Focus when necessary: Adaptive routing and collaborative grounding
  for training-free visual grounding.
\newblock \emph{arXiv preprint arXiv:2606.16158}, 2026.

\bibitem[Wu and Xie(2024)]{vstar2024}
Penghao Wu and Saining Xie.
\newblock V*: Guided visual search as a core mechanism in multimodal llms.
\newblock In \emph{CVPR}, 2024.

\bibitem[Yu et~al.(2025)Yu, Guan, and Gu]{zoomrefine2025}
Xuan Yu, Dayan Guan, and Yanfeng Gu.
\newblock Zoom-refine: Boosting high-resolution multimodal understanding via
  localized zoom and self-refinement.
\newblock \emph{arXiv preprint arXiv:2506.01663}, 2025.

\bibitem[Zhang et~al.(2025{\natexlab{a}})]{vicrop2025}
Jiarui Zhang et~al.
\newblock {MLLMs} know where to look: Training-free perception of small visual
  details with multimodal {LLMs}.
\newblock In \emph{ICLR}, 2025{\natexlab{a}}.

\bibitem[Zhang et~al.(2025{\natexlab{b}})Zhang, Gao, Zhang, Li,
  et~al.]{chainoffocus2025}
Xintong Zhang, Zhi Gao, Bofei Zhang, Pengxiang Li, et~al.
\newblock Adaptive chain-of-focus reasoning via dynamic visual search and
  zooming for efficient {VLMs}.
\newblock \emph{arXiv preprint arXiv:2505.15436}, 2025{\natexlab{b}}.

\bibitem[Zheng et~al.(2026)Zheng, Yang, Hong, Zhao, Xu, Yang, Shen, and
  Yu]{deepeyes2025}
Ziwei Zheng, Michael Yang, Jack Hong, Chenxiao Zhao, Guohai Xu, Le Yang, Chao
  Shen, and Xing Yu.
\newblock {DeepEyes}: Incentivizing ``thinking with images'' via reinforcement
  learning.
\newblock In \emph{ICLR}, 2026.

\end{thebibliography}
}

\clearpage
\appendix
\renewcommand{\topfraction}{0.95}\renewcommand{\bottomfraction}{0.8}\renewcommand{\textfraction}{0.05}
\setcounter{topnumber}{4}\setcounter{bottomnumber}{2}\setcounter{totalnumber}{6}\setcounter{dbltopnumber}{3}
\section*{Appendix}
This appendix gives what the main text cites but has no room for. \textbf{A} lists every prompt and the decoding settings. \textbf{B} gives the
statistics behind the main claims: the check without the tuning items, the four runs at 4B, the comparison with each baseline, and the
additional analyses quoted in the main text. \textbf{C} shows more examples. \textbf{D} gives SEE on Qwen3.5-27B~\cite{qwen2025} and every method's score on every
benchmark and backbone. \textbf{E} describes how each baseline was run, including HiDe on every backbone. \textbf{F}
breaks down SEE's inference time. At 4B, the \emph{comparison run} is the run of Tables~2--4, and the \emph{analysis run} that
of Table~5 and Figs.~1, 3 and~4.

\section{Prompts and implementation details}
\label{sec:supp:prompts}
\paragraph{Implementation details.} \label{sec:supp:impl}
All models run in \texttt{bfloat16} at native resolution up to $16.8$\,MP (larger images are resized to it); answers are
decoded greedily. \emph{\texttt{where}}: $K{=}3$ replies at temperature $0.8$, top-$p$ $0.95$, at most $256$ tokens, boxes
on a $0$--$1000$ grid; a reply without a box is dropped. \emph{\texttt{Zoom}}: the region is widened by $30\%$ of its size
on each side, clipped to the image, cropped, and enlarged (Lanczos) when its shorter side is below $448$ pixels.
\emph{\texttt{Read}}: greedy, at most $512$ tokens ($1024$ on InfographicVQA~\cite{infographicvqa2021}; $2048$ changes ANLS by $+0.2$), placed before
the question after ``OCR of the image:''. \emph{Vote}: open-ended answers are compared after lower-casing and removing
punctuation; $\tau{=}0.9$. An uncertain question costs $K$ batched region requests, one transcription and at most $K{+}2$
answers. \emph{Thinking mode}: each view first thinks (at most $1024$ tokens), and the reasoning goes before the question as
``Step-by-step analysis of the image''. GLM-4.1V's plain mode starts from an empty reasoning block, and its
\texttt{<|begin\_of\_box|>} answer tags are removed. \emph{Runs}: SEE ran four times at 4B (Table~\ref{tab:supp:runs}).
Tables~2--4 use the comparison run (mean $84.6$), which also evaluates the controls; Table~5 and
Figs.~1, 3 and~4 use the analysis run ($84.4$); the other two reach $85.4$ and $85.1$. Thinking mode has its own runs.
\emph{Cost}: wall-clock time per question, per stage (gated methods count extra stages only when the gate passes), as a
multiple of the direct answer in the same run, so machines cancel.

\paragraph{Prompts.}
Every prompt, verbatim. Each request is a system message followed by a user message holding the image and the text;
thinking is disabled except in the thinking mode.

\paragraph{Answering a view, multiple choice} (system; the user message is the image and the question followed by
the lettered options, one per line):
{\footnotesize
\begin{verbatim}
You are a visual question answering assistant.
Answer ONLY with the single letter (A, B, C, or D)
corresponding to the correct option. Do not
explain, do not add any other text.
\end{verbatim}
}
\paragraph{Answering a view, open-ended} (system):
{\footnotesize
\begin{verbatim}
Answer the question using the image. Give only the
exact answer (a short span), no explanation.
\end{verbatim}
}
\paragraph{Where to look} (system, then user; the options are included for multiple choice):
{\footnotesize
\begin{verbatim}
You identify the minimal visual evidence needed to
answer a question about an image.
\end{verbatim}
}
{\footnotesize
\begin{verbatim}
Question: {Q} Options: A) ... B) ... Identify the
MINIMAL visual evidence needed to decide among
these options. Output ONLY JSON: {"regions":
[[x1,y1,x2,y2], ...], "reason": "..."}. Use as
many or as few regions as truly needed; use an
empty list [] if the whole image is required.
Coordinates are normalized 0-1000. Respond with
ONLY the JSON object and nothing else - no
explanation, no markdown.
\end{verbatim}
}
{\footnotesize
\begin{verbatim}
Question: {Q} Identify the MINIMAL visual evidence
needed to answer this question. Output ONLY JSON:
{"regions": [[x1,y1,x2,y2], ...], "reason":
"..."}. Use as many or as few regions as truly
needed; use an empty list [] if the whole image is
required. Coordinates are normalized 0-1000.
Respond with ONLY the JSON object.
\end{verbatim}
}
\paragraph{What the image says} (system, then user):
{\footnotesize
\begin{verbatim}
You read text in images.
\end{verbatim}
}
{\footnotesize
\begin{verbatim}
Transcribe ALL text in this image, preserving
reading order. Output only the text.
\end{verbatim}
}
\paragraph{The \texttt{Read} view} (user text after the image):
{\footnotesize
\begin{verbatim}
OCR of the image: {transcript}

{question with options}
\end{verbatim}
}
\paragraph{Self-consistency baseline~\cite{selfconsistency2022}} (system, then the question followed by this instruction; three chains sampled at
temperature $0.8$ and top-$p$ $0.95$, up to $512$ tokens each; each chain is then answered like the \texttt{Read} view,
with \texttt{Step-by-step analysis of the image:} and the chain in place of the transcript):
{\footnotesize
\begin{verbatim}
You are a careful visual assistant.
\end{verbatim}
}
{\footnotesize
\begin{verbatim}
Answer by reasoning about the visual evidence,
step by step:
1. Locate: say which part of the image holds the
information needed (e.g. 'top-left', 'the sign
above the door'), as precisely as you can.
2. Zoom in: describe that region closely, as if
looking at an enlarged view of it: objects,
colours, positions, sizes.
3. Read: transcribe exactly any text, numbers or
labels in that region.
4. Conclude: using only this evidence, reason to
the answer.
Write the four steps.
\end{verbatim}
}
\paragraph{Thinking mode} (each view is first answered with thinking enabled using the answering prompt above; the
reasoning before \texttt{</think>} is then placed in the user text, and the answer is read with thinking disabled):
{\footnotesize
\begin{verbatim}
Step-by-step analysis of the image: {reasoning}

{question with options}
\end{verbatim}
}

\section{Results without the tuning items, and significance}
\label{sec:supp:heldout}
$K$, the crop margin and the crop size were chosen on 60-item subsets of the evaluation sets at 4B (Sec.~4,
Protocol). Table~\ref{tab:supp:heldout} re-scores every cell with those items removed. The vote rule and
$\tau$ were compared on the full sets (Table~5; Fig.~\ref{fig:supp:strengths}), so this check covers the first three choices only. Confidence intervals are paired bootstrap 95\% intervals over
questions ($10{,}000$ resamples); $p$-values are exact McNemar tests on multiple choice and paired sign-flip
permutation tests ($10{,}000$ permutations) on ANLS.

\begin{table}[!htbp]
\centering
\caption{\textbf{Gains without the tuning items} (points; Qwen3.5, analysis run at 4B). $K$, the crop margin and the crop size
were chosen on $60$ items per benchmark; removing those items ($n$: questions left) changes no gain by more than $1.2$
points, and every small-object gain stays positive. The interval is a paired bootstrap and $p$ an exact McNemar (or
sign-flip) test, so near $p{=}0.05$ the two can disagree (2B V*Bench).}
\setlength{\tabcolsep}{3pt}
\resizebox{\columnwidth}{!}{%
\begin{tabular}{llcccccc}
\toprule
& & \multicolumn{2}{c}{\textbf{All items}} & \multicolumn{4}{c}{\textbf{Without tuning items}} \\
\cmidrule(lr){3-4}\cmidrule(lr){5-8}
\textbf{Size} & \textbf{Benchmark} & $\Delta$ & $p$ & $\Delta$ & 95\% CI & $p$ & $n$ \\
\midrule
0.8B & V*Bench & $+13.1$ & $<10^{-3}$ & $+13.7$ & $[+7.6, +19.8]$ & $<10^{-3}$ & 131 \\
0.8B & HR-Bench 4K & $+4.4$ & $<10^{-3}$ & $+3.9$ & $[+2.0, +5.9]$ & $<10^{-3}$ & 740 \\
0.8B & HR-Bench 8K & $+6.6$ & $<10^{-3}$ & $+6.6$ & $[+4.3, +8.9]$ & $<10^{-3}$ & 740 \\
0.8B & InfographicVQA & $+1.0$ & $0.440$ & $+1.3$ & $[-1.2, +3.9]$ & $0.330$ & 440 \\
2B & V*Bench & $+5.8$ & $0.003$ & $+4.6$ & $[+0.8, +9.2]$ & $0.070$ & 131 \\
2B & HR-Bench 4K & $+3.4$ & $<10^{-3}$ & $+3.4$ & $[+1.6, +5.1]$ & $<10^{-3}$ & 740 \\
2B & HR-Bench 8K & $+4.0$ & $<10^{-3}$ & $+4.1$ & $[+2.0, +6.1]$ & $<10^{-3}$ & 740 \\
2B & InfographicVQA & $-1.3$ & $0.260$ & $-0.9$ & $[-3.6, +1.7]$ & $0.479$ & 440 \\
4B & V*Bench & $+11.0$ & $<10^{-3}$ & $+12.2$ & $[+6.1, +19.1]$ & $<10^{-3}$ & 131 \\
4B & HR-Bench 4K & $+4.6$ & $<10^{-3}$ & $+4.3$ & $[+2.2, +6.5]$ & $<10^{-3}$ & 740 \\
4B & HR-Bench 8K & $+7.6$ & $<10^{-3}$ & $+7.4$ & $[+5.0, +10.0]$ & $<10^{-3}$ & 740 \\
4B & InfographicVQA & $+2.6$ & $0.016$ & $+3.0$ & $[+0.7, +5.2]$ & $0.013$ & 440 \\
9B & V*Bench & $+4.2$ & $0.039$ & $+4.6$ & $[+1.5, +8.4]$ & $0.031$ & 131 \\
9B & HR-Bench 4K & $+3.6$ & $<10^{-3}$ & $+3.5$ & $[+1.6, +5.4]$ & $<10^{-3}$ & 740 \\
9B & HR-Bench 8K & $+6.2$ & $<10^{-3}$ & $+6.8$ & $[+4.7, +8.9]$ & $<10^{-3}$ & 740 \\
9B & InfographicVQA & $+1.4$ & $0.102$ & $+1.8$ & $[+0.1, +3.5]$ & $0.038$ & 440 \\
\bottomrule
\end{tabular}}
\label{tab:supp:heldout}
\end{table}

\paragraph{Runs at 4B.} SEE ran four times at 4B with the same setting; the runs differ only in the sampling of the
region requests (Table~\ref{tab:supp:runs}). The analysis run, used for ablations, has the lowest mean.

\begin{table}[!htbp]
\centering
\caption{\textbf{SEE at 4B across four runs} (accuracy, ANLS on InfographicVQA; gain over the direct answer in brackets). The
runs differ only in the sampled region requests, and the gain moves little between them (standard deviation at most
$0.7$ points); the analysis run, used for the ablations, is the weakest of the four.}
\setlength{\tabcolsep}{2.5pt}
\resizebox{\columnwidth}{!}{%
\begin{tabular}{lccccc}
\toprule
\textbf{Run} & \textbf{V*Bench} & \textbf{HR-4K} & \textbf{HR-8K} & \textbf{InfoVQA} & \textbf{Mean} \\
\midrule
analysis run & 89.5 ($+11.0$) & 82.3 ($+4.6$) & 81.0 ($+7.6$) & 84.7 ($+2.6$) & 84.4 \\
comparison run & 90.1 ($+11.5$) & 82.6 ($+5.0$) & 81.4 ($+8.0$) & 84.2 ($+2.2$) & 84.6 \\
seed 2 & 90.6 ($+12.0$) & 83.4 ($+5.8$) & 81.9 ($+8.5$) & 85.6 ($+3.6$) & 85.4 \\
seed 3 & 90.6 ($+12.0$) & 83.5 ($+5.9$) & 81.0 ($+7.6$) & 85.4 ($+3.4$) & 85.1 \\
\midrule
mean $\pm$ SD of the gain & $+11.7\pm0.5$ & $+5.3\pm0.6$ & $+7.9\pm0.4$ & $+3.0\pm0.7$ & \\
\bottomrule
\end{tabular}}
\label{tab:supp:runs}
\end{table}

The same paired bootstrap compares SEE with each baseline of Table~2, question by question on the same
items (Table~\ref{tab:supp:boot}); an aggregate is the mean of its cells, each resampled within itself. SEE's
small-object average is significantly above every baseline's; overall it is on par with self-consistency, which is
significantly stronger on InfographicVQA.

\begin{table}[!htbp]
\centering
\caption{\textbf{SEE against each baseline} (points; accuracy, ANLS on InfographicVQA; plain mode, full sets, comparison run at
4B). $\Delta$: SEE minus the baseline, averaged over the row's cells (per-backbone rows: the four benchmarks), with a
95\% CI and two-sided $p$ from a paired bootstrap over questions, stratified by cell ($10{,}000$ resamples). On small
objects, SEE is significantly ahead of every baseline.}
\setlength{\tabcolsep}{3pt}
\resizebox{\columnwidth}{!}{%
\begin{tabular}{llccc}
\toprule
\textbf{Baseline} & \textbf{Cells} & $\Delta$ & 95\% CI & $p$ \\
\midrule
Self-consistency & Small$^\circ$ (15 cells) & $+2.3$ & $[+1.4, +3.2]$ & $<10^{-3}$ \\
 & InfoVQA (5 cells) & $-5.1$ & $[-6.4, -3.9]$ & $<10^{-3}$ \\
 & Overall$^\circ$ (20 cells) & $+0.4$ & $[-0.3, +1.2]$ & $0.253$ \\
 & Q3.5-0.8B & $+0.3$ & $[-1.8, +2.5]$ & $0.758$ \\
 & Q3.5-2B & $0.0$ & $[-1.7, +1.6]$ & $0.951$ \\
 & Q3.5-4B & $+1.1$ & $[-0.6, +2.8]$ & $0.222$ \\
 & Q3.5-9B & $0.0$ & $[-1.4, +1.5]$ & $0.962$ \\
 & GLM-4.1V & $+0.8$ & $[-0.5, +2.2]$ & $0.235$ \\
\midrule
HiDe & Small$^\circ$ (15 cells) & $+1.8$ & $[+0.7, +3.0]$ & $0.001$ \\
 & InfoVQA (5 cells) & $+0.3$ & $[-1.1, +1.7]$ & $0.711$ \\
 & Overall$^\circ$ (20 cells) & $+1.4$ & $[+0.5, +2.3]$ & $0.001$ \\
 & Q3.5-0.8B & $+2.3$ & $[+0.1, +4.7]$ & $0.042$ \\
 & Q3.5-2B & $+2.6$ & $[+0.2, +4.9]$ & $0.029$ \\
 & Q3.5-4B & $+8.0$ & $[+5.8, +10.3]$ & $<10^{-3}$ \\
 & Q3.5-9B & $-3.9$ & $[-5.4, -2.5]$ & $<10^{-3}$ \\
 & GLM-4.1V & $-1.8$ & $[-3.5, -0.1]$ & $0.035$ \\
\midrule
Zoom-Refine & Small$^\circ$ (15 cells) & $+3.5$ & $[+2.5, +4.4]$ & $<10^{-3}$ \\
 & InfoVQA (5 cells) & $+4.8$ & $[+3.4, +6.2]$ & $<10^{-3}$ \\
 & Overall$^\circ$ (20 cells) & $+3.8$ & $[+3.0, +4.6]$ & $<10^{-3}$ \\
 & Q3.5-0.8B & $+4.3$ & $[+2.4, +6.1]$ & $<10^{-3}$ \\
 & Q3.5-2B & $+9.6$ & $[+7.7, +11.5]$ & $<10^{-3}$ \\
 & Q3.5-4B & $+4.2$ & $[+2.7, +5.8]$ & $<10^{-3}$ \\
 & Q3.5-9B & $+1.4$ & $[-0.1, +3.0]$ & $0.073$ \\
 & GLM-4.1V & $-0.5$ & $[-2.2, +1.2]$ & $0.567$ \\
\midrule
DyFo & Small$^\circ$ (15 cells) & $+6.8$ & $[+5.8, +7.8]$ & $<10^{-3}$ \\
 & InfoVQA (5 cells) & $+4.4$ & $[+3.0, +5.8]$ & $<10^{-3}$ \\
 & Overall$^\circ$ (20 cells) & $+6.2$ & $[+5.4, +7.0]$ & $<10^{-3}$ \\
 & Q3.5-0.8B & $+6.4$ & $[+4.4, +8.5]$ & $<10^{-3}$ \\
 & Q3.5-2B & $+5.4$ & $[+3.5, +7.2]$ & $<10^{-3}$ \\
 & Q3.5-4B & $+13.6$ & $[+11.6, +15.7]$ & $<10^{-3}$ \\
 & Q3.5-9B & $+5.3$ & $[+3.7, +6.9]$ & $<10^{-3}$ \\
 & GLM-4.1V & $+0.3$ & $[-1.4, +2.0]$ & $0.727$ \\
\midrule
ZoomEye & Small$^\circ$ (15 cells) & $+5.4$ & $[+3.8, +7.1]$ & $<10^{-3}$ \\
 & InfoVQA (5 cells) & $+2.7$ & $[+1.5, +3.9]$ & $<10^{-3}$ \\
 & Overall$^\circ$ (20 cells) & $+4.7$ & $[+3.5, +6.0]$ & $<10^{-3}$ \\
 & Q3.5-0.8B & $+7.7$ & $[+4.4, +10.9]$ & $<10^{-3}$ \\
 & Q3.5-2B & $+6.2$ & $[+3.1, +9.3]$ & $<10^{-3}$ \\
 & Q3.5-4B & $+6.7$ & $[+3.7, +9.7]$ & $<10^{-3}$ \\
 & Q3.5-9B & $+0.2$ & $[-2.3, +2.8]$ & $0.871$ \\
 & GLM-4.1V & $+2.9$ & $[+0.6, +5.3]$ & $0.010$ \\
\bottomrule
\end{tabular}}
\label{tab:supp:boot}
\end{table}

\paragraph{Additional analyses.} Table~\ref{tab:supp:extra} collects the numbers the main text quotes without a table of
its own, and Fig.~\ref{fig:supp:strengths} adds three views of why SEE works: it lifts every model size, the agreement of
its zooms signals a right answer, and the gate halves the time at no loss.

\begin{figure*}[!t]
\centering
\setlength{\tabcolsep}{1pt}
\begin{tabular}{ccc}
\includegraphics[width=0.32\textwidth]{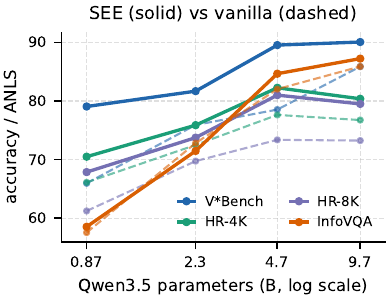} &
\includegraphics[width=0.32\textwidth]{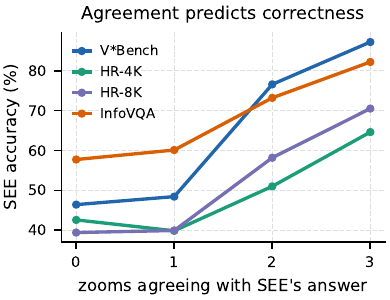} &
\includegraphics[width=0.32\textwidth]{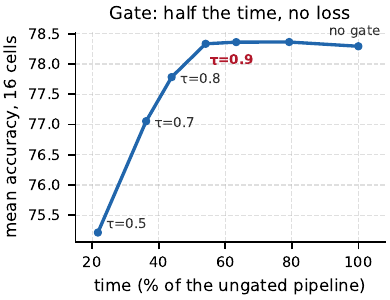} \\
\end{tabular}
\caption{\textbf{More on why SEE works} (Qwen3.5, plain mode, full sets). \emph{Left:} accuracy against model size, with SEE
(solid) and the direct answer (dashed); SEE lifts every size on small objects, and 4B with SEE beats the 9B direct
answer. \emph{Middle:} SEE's accuracy by how many of its $3$ zooms agree with its final answer; agreement signals a
right answer, which the vote rewards. \emph{Right:} mean accuracy of the $16$ Qwen3.5 cells against time for different
gates $\tau$; $\tau{=}0.9$ keeps the accuracy at half the time, and any $\tau\in[0.9,0.99]$ moves the mean by at most
$0.1$.}
\label{fig:supp:strengths}
\end{figure*}

\begin{table}[!htbp]
\centering
\caption{\textbf{Additional analyses quoted in the main text} (Qwen3.5-4B unless stated; plain mode, full sets). Each row
gives the number behind one statement in the main text.}
\setlength{\tabcolsep}{3pt}
\footnotesize
\begin{tabular}{@{}p{0.62\columnwidth}@{\hspace{4pt}}p{0.34\columnwidth}@{}}
\toprule
\textbf{Analysis} & \textbf{Result} \\
\midrule
Elicitation gap (V*Bench; medoid of 5 box requests, hit: IoU $\ge 0.5$) & hits on $20/41$ wrong ($49\%$), $118/150$ right ($79\%$) \\
IoU of that box predicts that a zoom fixes the error & AUROC $0.78$ ($n{=}41$) \\
Self-check ``does this crop show what the question asks?'' (pilots) & AUROC $0.83$--$0.88$; $\le 2$ answers changed \\
Most confident view instead of the vote (every size) & small objects $+1.4$ to $+2.1$; InfoVQA $-0.4$ to $-2.4$ \\
The $K$ zooms share one vote (weight $1/K$) instead of one each & mean $81.9$ vs.\ $84.4$ ($-2.5$) \\
Calibration error of zoomed views vs.\ direct answer (small objects) & ECE $0.05$--$0.09$ vs.\ $0.05$--$0.10$ \\
One temperature per view type (held-out questions) & accuracy $+0.3$ \\
Direct + reading view only, InfoVQA: own / empty / other image's transcript & $+2.2$ / $-0.2$ / $-1.1$ ANLS \\
Zoomed views of InfoVQA pages: mean confidence vs.\ share right & $69\%$ vs.\ $55\%$ \\
\bottomrule
\end{tabular}

\vspace{6pt}
\setlength{\tabcolsep}{3.5pt}
\begin{tabular}{lccccc}
\toprule
\textbf{Gate stop rate (\%)} & \textbf{0.8B} & \textbf{2B} & \textbf{4B} & \textbf{9B} & \textbf{GLM} \\
\midrule
V*Bench & 30 & 56 & 43 & 60 & 78 \\
HR-Bench 4K & 46 & 57 & 56 & 60 & 71 \\
HR-Bench 8K & 39 & 52 & 50 & 53 & 64 \\
InfographicVQA & 34 & 35 & 53 & 60 & 86 \\
\midrule
\multicolumn{6}{l}{thinking mode, 4B: $87$--$92\%$ over the four benchmarks} \\
\bottomrule
\end{tabular}

\vspace{6pt}
\begin{tabular}{lccccc}
\toprule
\textbf{HR-Bench gain} & \textbf{0.8B} & \textbf{2B} & \textbf{4B} & \textbf{9B} & \textbf{GLM} \\
\midrule
4K, single-object & $+8.3$ & $+5.5$ & $+7.3$ & $+5.0$ & $+6.8$ \\
4K, cross-object & $+0.5$ & $+1.3$ & $+2.8$ & $+2.3$ & $-2.3$ \\
8K, single-object & $+11.0$ & $+7.3$ & $+12.0$ & $+5.3$ & $+10.0$ \\
8K, cross-object & $+2.3$ & $+0.8$ & $+4.0$ & $+7.3$ & $0.0$ \\
\bottomrule
\end{tabular}
\label{tab:supp:extra}
\end{table}

\section{More examples}
\begin{figure}[!htbp]
\centering
\includegraphics[width=\columnwidth]{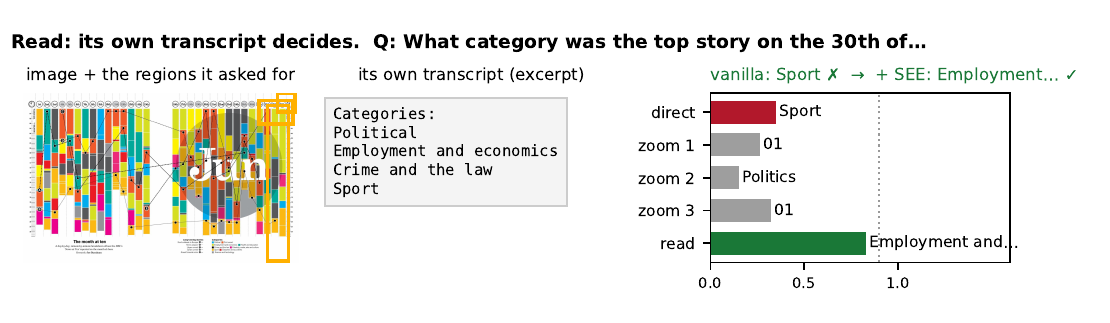}
\caption{\textbf{Reading decides} (InfographicVQA, Qwen3.5-4B). The three zooms disagree with one another, while the model's
own transcript contains the answer, so the reading view outvotes the direct answer.}
\label{fig:supp:ex_read}
\end{figure}

\begin{figure}[!htbp]
\centering
\includegraphics[width=\columnwidth]{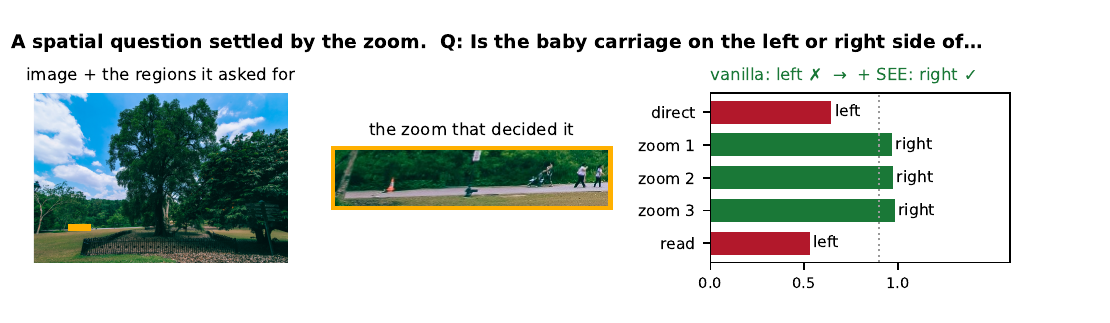}
\caption{\textbf{A spatial question settled by the zoom} (V*Bench, Qwen3.5-4B). All three zooms put the baby carriage to the
right of the cone and overturn an unsure direct answer.}
\label{fig:supp:ex_carriage}
\end{figure}

\begin{figure}[!htbp]
\centering
\includegraphics[width=\columnwidth]{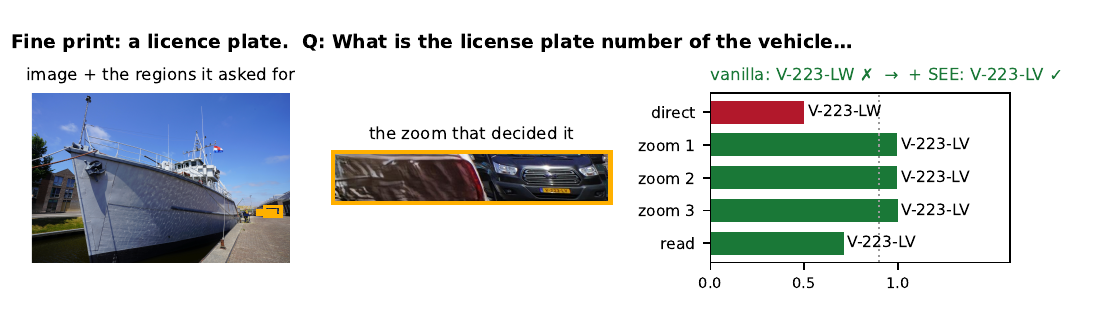}
\caption{\textbf{Fine print} (HR-Bench 8K, Qwen3.5-4B). The direct answer misreads one character of a licence plate (LW instead
of LV); every zoom reads it correctly.}
\label{fig:supp:ex_plate}
\end{figure}

\begin{figure}[!htbp]
\centering
\includegraphics[width=\columnwidth]{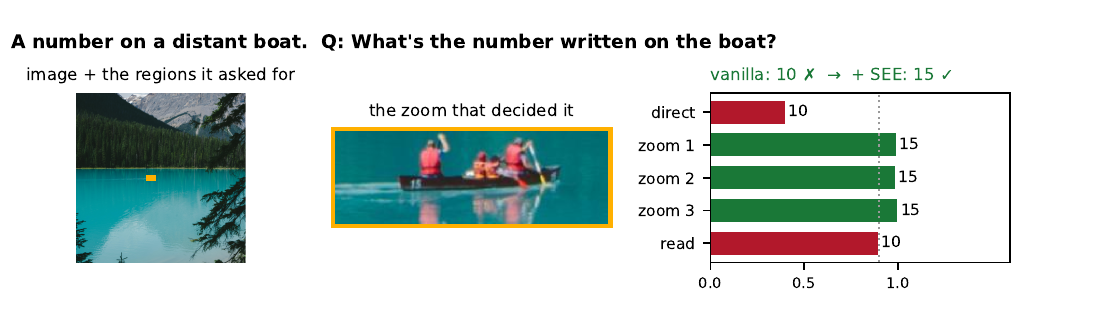}
\caption{\textbf{A number on a distant boat} (HR-Bench 4K, Qwen3.5-4B). All three zooms read ``15'', while the direct answer
and the transcript both say ``10''; the zooms win the vote.}
\label{fig:supp:ex_boat}
\end{figure}

\begin{figure}[!htbp]
\centering
\includegraphics[width=\columnwidth]{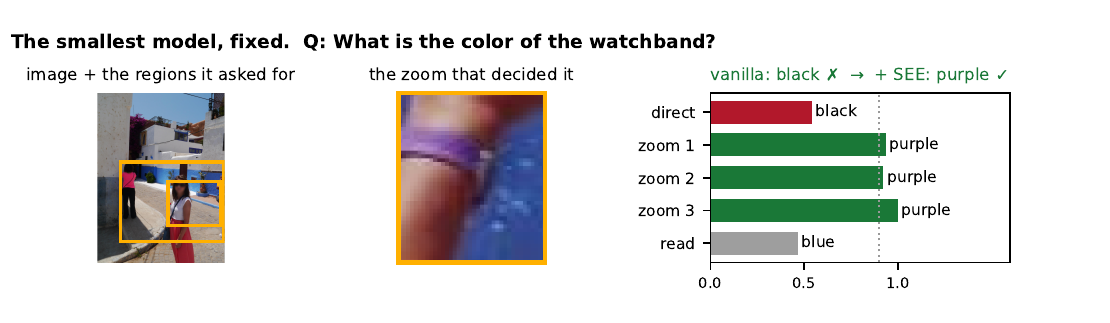}
\caption{\textbf{The smallest model, fixed} (V*Bench, Qwen3.5-0.8B). On the whole image the 0.8B model calls the watchband
black; on all three of its own zooms it answers purple, confidently and correctly.}
\label{fig:supp:ex_watch}
\end{figure}

\begin{figure}[!htbp]
\centering
\includegraphics[width=\columnwidth]{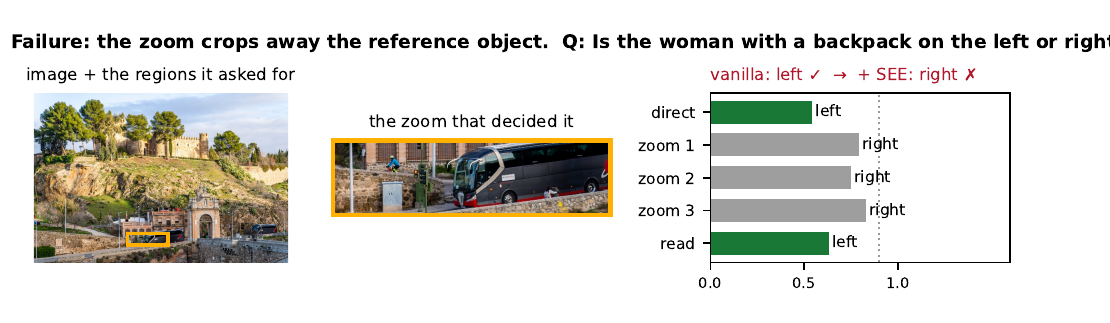}
\caption{\textbf{A failure: the crop drops the reference object} (V*Bench, Qwen3.5-4B). The question asks where one object is
relative to another, but the zoom keeps only one of them, and three confident zooms flip a correct direct answer.}
\label{fig:supp:ex_fail_crop}
\end{figure}

\begin{figure}[!htbp]
\centering
\includegraphics[width=\columnwidth]{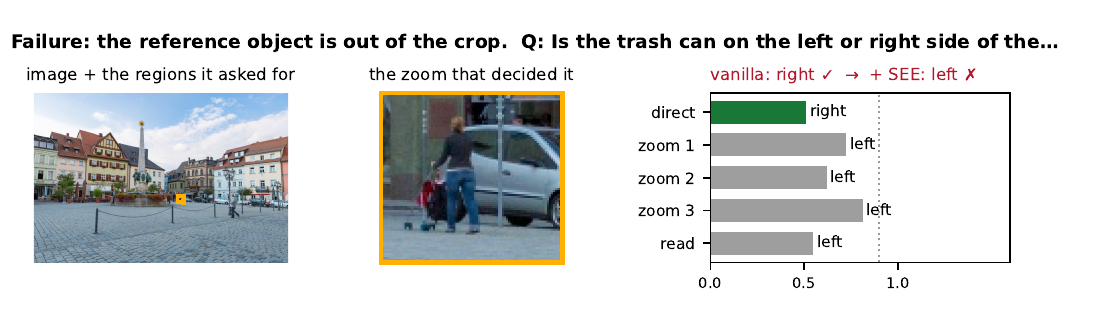}
\caption{\textbf{The same failure} (V*Bench, Qwen3.5-4B). Again the zoom keeps only one of the two objects the question
relates, and the zooms outvote a correct direct answer.}
\label{fig:supp:ex_fail_crop2}
\end{figure}

Figs.~\ref{fig:supp:ex_read}--\ref{fig:supp:ex_fail_crop2} show single questions from the analysis runs larger, in the same format as Fig.~6. Left: the image and the regions the model asked for; middle: the evidence that decided the answer; right: every view's confidence $p_v$, green for the right answer, red for vanilla's wrong answer, grey for any other wrong answer; dotted line: the skip gate $\tau{=}0.9$. The cases were mined automatically from the logs by category (a zoom fixes it, reading fixes it, the vote rescues it, failure) and checked by eye; the scripts and the full set of $131$ examples will be released with the code.

\section{Detailed results on every backbone}
\label{sec:supp:full}
Table~\ref{tab:full-results} gives the per-benchmark scores behind the averages of Table~2. On the Qwen3.5 backbones, SEE's gain over the direct answer is significant on every small-target benchmark
(exact McNemar test, $p<0.05$); on GLM-4.1V-9B~\cite{glm41v} it is significant on V*Bench~\cite{vstar2024} and HR-Bench 8K~\cite{hrbench2024} ($p=0.04$, $p<0.001$)
but not on HR-Bench 4K ($p=0.05$) or InfographicVQA. On InfographicVQA the gain is significant only for
Qwen3.5-4B. Table~\ref{tab:supp:heldout} repeats the tests with the tuning items removed, and
Table~\ref{tab:supp:recdmg} breaks SEE's recovery and damage down by benchmark. Table~\ref{tab:supp:text} gives the change in ANLS on InfographicVQA for every method, and
Table~\ref{tab:supp:think} the thinking-mode results behind Table~6, and Table~\ref{tab:supp:27b} SEE on Qwen3.5-27B.

\begin{table*}[!t]
  \centering
  \caption{
    \textbf{Every method on every backbone and benchmark} (accuracy, ANLS on InfographicVQA; \%). Both blocks share the method and dataset order; the reading guide explains the abbreviations and marks.}
  \label{tab:full-results}
  \begingroup
  \fontsize{8}{9.6}\selectfont
  \renewcommand{\arraystretch}{1.15}
  \setlength{\tabcolsep}{0pt}
  \edef\cvprMethodW{\the\dimexpr\textwidth*13/100\relax}
  \edef\cvprScoreW{\the\dimexpr(\textwidth-\cvprMethodW)/16\relax}
  \edef\cvprVstarW{\the\dimexpr\cvprScoreW*8/10\relax}
  \edef\cvprHRW{\the\dimexpr\cvprScoreW*39/40\relax}
  \edef\cvprInfoW{\the\dimexpr\cvprScoreW*5/4\relax}
  \edef\cvprLowerW{\the\dimexpr\cvprMethodW+\cvprScoreW*4\relax}
  \edef\cvprLegendW{\the\dimexpr\textwidth-\cvprLowerW\relax}

  \begin{tabular}{
    @{} >{\raggedright\arraybackslash}p{\cvprMethodW}
    *{4}{
      >{\raggedleft\arraybackslash}p{\cvprVstarW}
      *{2}{>{\raggedleft\arraybackslash}p{\cvprHRW}}
      >{\raggedleft\arraybackslash}p{\cvprInfoW}
    } @{}
  }
    \toprule
    Method & \multicolumn{4}{c}{\textbf{Q3.5-0.8B}} & \multicolumn{4}{c}{\textbf{Q3.5-2B}} & \multicolumn{4}{c}{\textbf{Q3.5-4B}} & \multicolumn{4}{c}{\textbf{Q3.5-9B}} \\
    \cmidrule(lr{3pt}){2-5}
    \cmidrule(lr{3pt}){6-9}
    \cmidrule(lr{3pt}){10-13}
    \cmidrule(lr{3pt}){14-17}
      & V* & HR-4K & HR-8K & InfoVQA & V* & HR-4K & HR-8K & InfoVQA & V* & HR-4K & HR-8K & InfoVQA & V* & HR-4K & HR-8K & InfoVQA \\
    \midrule
    Direct
      & 66.0 & 66.1 & 61.3 & 57.6
      & 75.9 & 72.5 & 69.8 & 72.8
      & 78.5 & 77.6 & 73.4 & 82.0
      & 85.9 & 76.7 & 73.3 & 85.9 \\
    Self-consistency
      & 74.3 & 68.6 & 64.5 & \cvprTableBest{67.2}
      & 74.9 & \cvprTableSecond{74.9} & \cvprTableSecond{71.9} & \cvprTableBest{81.3}
      & \cvprTableSecond{82.2} & \cvprTableBest{84.9} & \cvprTableSecond{80.0} & \cvprTableBest{87.0}
      & 88.5 & \cvprTableSecond{81.1} & 77.8 & \cvprTableBest{89.7} \\
    Zoom-Refine
      & 74.3 & 69.2 & 65.5 & 49.8
      & 71.2 & 68.1 & 65.1 & 60.1
      & 81.7 & 80.5 & 78.1 & 81.1
      & 83.2 & \cvprTableSecond{81.1} & \cvprTableSecond{80.0} & \cvprTableSecond{87.2} \\
    ZoomEye
      & 66.0 & 69.0 & 59.0 & 49.0
      & 72.8 & 67.0 & 72.0 & 69.5
      & 80.6 & 80.0 & 70.0 & 82.8
      & \cvprTableBest{93.7} & 77.0 & 79.0 & 86.9 \\
    HiDe
      & 74.3 & \cvprTableBest{70.8} & \cvprTableSecond{66.1} & 55.6
      & 73.8 & 73.2 & 71.2 & \cvprTableSecond{74.2}
      & 73.3 & 76.0 & 73.1 & 83.7
      & \cvprTableSecond{92.7} & \cvprTableBest{87.4} & \cvprTableBest{85.8} & 86.9 \\
    DyFo
      & \cvprTableSecond{77.0} & 57.1 & 55.1 & \cvprTableSecond{61.1}
      & \cvprTableBest{84.8} & 66.6 & 63.6 & 66.2
      & \cvprTableSecond{82.2} & 65.1 & 60.5 & 75.8
      & 84.3 & 77.0 & 74.2 & 80.6 \\
    \midrule
    \rowcolor{cvprOursGray}
    \textbf{SEE (Ours)}
      & \cvprTableBest{79.1} & \cvprTableSecond{70.5} & \cvprTableBest{67.9} & 58.6
      & \cvprTableSecond{81.7} & \cvprTableBest{75.9} & \cvprTableBest{73.8} & 71.5
      & \cvprTableBest{90.1} & \cvprTableSecond{82.6} & \cvprTableBest{81.4} & \cvprTableSecond{84.2}
      & 90.1 & 80.4 & 79.5 & \cvprTableSecond{87.2} \\
    \bottomrule
  \end{tabular}
  \par\vspace{8pt}

  \noindent
  \begin{minipage}[c]{\cvprLowerW}
    \vspace{0pt}
    \begin{tabular}{
      @{} >{\raggedright\arraybackslash}p{\cvprMethodW}
      *{1}{
        >{\raggedleft\arraybackslash}p{\cvprVstarW}
        *{2}{>{\raggedleft\arraybackslash}p{\cvprHRW}}
        >{\raggedleft\arraybackslash}p{\cvprInfoW}
      } @{}
    }
      \toprule
      Method & \multicolumn{4}{c}{\textbf{GLM-4.1V}} \\
      \cmidrule(lr{3pt}){2-5}
        & V* & HR-4K & HR-8K & InfoVQA \\
      \midrule
      Direct
        & 75.4 & 71.5 & 62.3 & 83.7 \\
      Self-consistency
        & 77.0 & 72.9 & 65.5 & \cvprTableBest{85.8} \\
      Zoom-Refine
        & 74.3 & \cvprTableBest{78.0} & \cvprTableBest{70.8} & 83.3 \\
      ZoomEye
        & 75.9 & 70.0 & 60.0 & \cvprTableSecond{83.9} \\
      HiDe
        & \cvprTableBest{80.1} & \cvprTableSecond{77.9} & \cvprTableSecond{70.1} & 83.5 \\
      DyFo
        & \cvprTableBest{80.1} & 73.9 & 69.6 & 79.6 \\
      \midrule
      \rowcolor{cvprOursGray}
      \textbf{SEE (Ours)}
        & \cvprTableSecond{79.6} & 73.8 & 67.3 & 83.8 \\
      \bottomrule
    \end{tabular}
  \end{minipage}%
  \begin{minipage}[c]{\cvprLegendW}
    \vspace{0pt}
    \hspace*{10pt}%
    \begin{minipage}[t]{\dimexpr\linewidth-10pt\relax}
      \vspace{4pt}
      \raggedright
      \fontsize{8}{10}\selectfont
      \textbf{Reading guide}\\[3pt]
      Q3.5: Qwen3.5. GLM-4.1V: GLM-4.1V-9B-Thinking, plain mode.\\[2pt]
      V*: V*Bench; HR-4K/8K: HR-Bench; InfoVQA: InfographicVQA.\\[2pt]
      \textbf{Bold}\,/\,\underline{underline}: best\,/\,second-best result in each model--dataset column; ties share the
      mark.\\[2pt]
      SEE on Q3.5-4B is the comparison run. Zoom-Refine and ZoomEye are our re-implementations with the authors'
      published settings; ZoomEye on Q3.5-9B InfoVQA covers $494$ of $500$ questions ($6$ exceeded GPU memory).\\[2pt]
    \end{minipage}
  \end{minipage}
  \endgroup
\end{table*}

\begin{table}[!htbp]
\centering
\caption{\textbf{SEE's recovery\,/\,damage per benchmark} (\%; plain mode, full sets, comparison run at 4B). Recovery is the
share of the direct answer's errors that SEE fixes, damage the share of its right answers that SEE breaks (on
InfographicVQA, right means ANLS $\ge 0.5$). On small objects SEE breaks at most $3\%$ of right answers on every
Qwen3.5 model ($5\%$ on GLM-4.1V) while fixing up to $59\%$ of errors.}
\setlength{\tabcolsep}{4pt}
\resizebox{\columnwidth}{!}{%
\begin{tabular}{lcccc|c}
\toprule
\textbf{Benchmark} & \textbf{0.8B} & \textbf{2B} & \textbf{4B} & \textbf{9B} & \textbf{GLM-9B} \\
\midrule
V*Bench        & $42\,/\,2$ & $26\,/\,1$ & $59\,/\,1$ & $37\,/\,1$ & $21\,/\,1$ \\
HR-Bench 4K    & $18\,/\,2$ & $18\,/\,2$ & $32\,/\,3$ & $25\,/\,3$ & $21\,/\,5$ \\
HR-Bench 8K    & $22\,/\,3$ & $19\,/\,3$ & $36\,/\,2$ & $28\,/\,2$ & $22\,/\,5$ \\
\midrule
InfographicVQA & $12\,/\,7$ & $15\,/\,6$ & $31\,/\,4$ & $18\,/\,2$ & $11\,/\,2$ \\
\bottomrule
\end{tabular}}
\label{tab:supp:recdmg}
\end{table}

\begin{table}[!htbp]
\centering
\caption{\textbf{Dense text} (InfographicVQA; change in ANLS against the direct answer; plain mode, full sets, comparison run
at 4B; GLM: GLM-4.1V). Crop- and search-based methods lose up to $12.7$ ANLS, mostly on the small models (DyFo gains
only at 0.8B), while SEE, the only method that also reads, stays within $1.3$. Self-consistency reasons over the whole
page instead of cropping it.}
\setlength{\tabcolsep}{3.5pt}
\resizebox{\columnwidth}{!}{%
\begin{tabular}{lccccc}
\toprule
\textbf{Method} & \textbf{0.8B} & \textbf{2B} & \textbf{4B} & \textbf{9B} & \textbf{GLM} \\
\midrule
Zoom-Refine & \gdn{$-7.8$} & \gdn{$-12.7$} & \gdn{$-0.9$} & \gup{$+1.3$} & \gdn{$-0.4$} \\
ZoomEye     & \gdn{$-8.6$} & \gdn{$-3.3$} & \gup{$+0.8$} & \gup{$+1.0$} & \gup{$+0.2$} \\
DyFo        & \gup{$+3.5$} & \gdn{$-6.6$} & \gdn{$-6.2$} & \gdn{$-5.3$} & \gdn{$-4.1$} \\
HiDe        & \gdn{$-2.0$} & \gup{$+1.4$} & \gup{$+1.7$} & \gup{$+1.0$} & \gdn{$-0.2$} \\
\rowcolor{gray!15}
\textbf{SEE} (zoom + read) & \gup{$+1.0$} & \gdn{$-1.3$} & \gup{$+2.2$} & \gup{$+1.3$} & \gup{$+0.1$} \\
\midrule
Self-consistency & \gup{$+9.6$} & \gup{$+8.5$} & \gup{$+5.0$} & \gup{$+3.8$} & \gup{$+2.1$} \\
\bottomrule
\end{tabular}}
\label{tab:supp:text}
\end{table}

\begin{table}[!htbp]
\centering
\caption{\textbf{Thinking mode per benchmark} (full sets; accuracy, ANLS on InfographicVQA). \emph{Thinking}: the model reasons
before answering; \emph{+ self-consistency}: three extra thinking chains under SEE's gate and vote; \emph{+ SEE}:
every view thinks. The gate stops $87$--$92\%$ of the questions at 4B and $96$--$100\%$ on GLM-4.1V. Exact McNemar $p$
of SEE vs.\ thinking alone on V*Bench, HR-4K and HR-8K: 0.8B 0.267, 0.014, 0.049; 2B 0.146, 0.135, 0.001; 4B 0.219,
${<}0.001$, ${<}0.001$; 9B 0.125, 0.052, 0.001.}
\setlength{\tabcolsep}{3pt}
\resizebox{\columnwidth}{!}{%
\begin{tabular}{llcccc}
\toprule
\textbf{Backbone} & \textbf{Mode} & \textbf{V*} & \textbf{HR-4K} & \textbf{HR-8K} & \textbf{InfoVQA} \\
\midrule
Q3.5-0.8B & thinking & 72.8 & 71.6 & 68.9 & 71.3 \\
 & + self-consistency & 75.4 & 73.6 & 70.0 & 74.2 \\
 & + SEE & 75.4 & 73.6 & 70.8 & 71.9 \\
\midrule
Q3.5-2B & thinking & 83.2 & 76.5 & 73.9 & 82.5 \\
 & + self-consistency & 83.2 & 78.8 & 76.1 & 85.1 \\
 & + SEE & 86.4 & 77.9 & 77.1 & 82.5 \\
\midrule
Q3.5-4B & thinking & 85.3 & 87.1 & 82.2 & 92.4 \\
 & + self-consistency & 85.9 & 88.1 & 84.0 & 93.0 \\
 & + SEE & 87.4 & 89.6 & 84.9 & 92.9 \\
\midrule
Q3.5-9B & thinking & 88.0 & 87.9 & 83.5 & 93.0 \\
 & + self-consistency & 88.5 & 88.9 & 83.6 & 93.2 \\
 & + SEE & 90.0 & 89.1 & 85.9 & 92.7 \\
\midrule
GLM-4.1V & thinking & 76.4 & 74.0 & 66.2 & 87.4 \\
 & + self-consistency & 76.4 & 74.2 & 66.0 & 87.7 \\
 & + SEE & 76.4 & 74.1 & 65.8 & 86.9 \\
\bottomrule
\end{tabular}}
\label{tab:supp:think}
\end{table}

\paragraph{Scaling to Qwen3.5-27B.} To see how SEE behaves beyond 9B, we ran it unchanged on Qwen3.5-27B (plain mode,
the 4B setting, one run; Table~\ref{tab:supp:27b}). The gain narrows with scale, from $+8.2$ points on small objects at 4B
and $+4.7$ at 9B to $+2.7$ at 27B, but stays positive on every benchmark; only the HR-Bench 8K gain is significant on its
own. SEE remains safe: on small objects it fixes $26.9\%$ of the direct answer's errors and breaks $3.1\%$ of its right
answers, in the range of the other backbones. The gate does not saturate (it stops $46$--$66\%$ of the questions, as at
9B), so the smaller gain reflects fewer errors left to fix rather than SEE acting less often. The one-model-size pattern
of the main text extends one step: on small objects, 9B with SEE ($83.3$) matches the 27B direct answer ($83.1$), and 4B
with SEE ($84.7$) beats it, in all four 4B runs. On InfographicVQA the 27B direct answer stays well ahead of 4B with SEE
($89.6$ vs.\ $84.2$). Baselines and thinking mode were not run at 27B.

\begin{table}[!htbp]
\centering
\caption{\textbf{SEE on Qwen3.5-27B} (plain mode, full sets, one run; the 4B setting unchanged; accuracy, ANLS on
InfographicVQA). SEE gains on every benchmark, most clearly on HR-Bench 8K ($+3.2$, $p{=}0.002$). $p$: exact McNemar test
of SEE vs.\ the direct answer (paired sign-flip test on ANLS); \emph{Gate stops}: share of questions answered directly.}
\setlength{\tabcolsep}{3.5pt}
\resizebox{\columnwidth}{!}{%
\begin{tabular}{lccccc}
\toprule
\textbf{Benchmark} & \textbf{Direct} & \textbf{SEE} & \textbf{Gain} & $p$ & \textbf{Gate stops (\%)} \\
\midrule
V*Bench        & 89.0 & 92.1 & $+3.1$ & 0.18  & 46.1 \\
HR-Bench 4K    & 82.0 & 83.8 & $+1.8$ & 0.09  & 54.1 \\
HR-Bench 8K    & 78.4 & 81.6 & $+3.2$ & 0.002 & 48.5 \\
InfographicVQA & 89.6 & 90.4 & $+0.9$ & 0.33  & 65.6 \\
\midrule
Small-object mean & 83.1 & 85.8 & $+2.7$ & & \\
Overall mean      & 84.8 & 87.0 & $+2.2$ & & \\
\bottomrule
\end{tabular}}
\label{tab:supp:27b}
\end{table}

\begin{figure*}[!t]
\centering
\includegraphics[width=\textwidth]{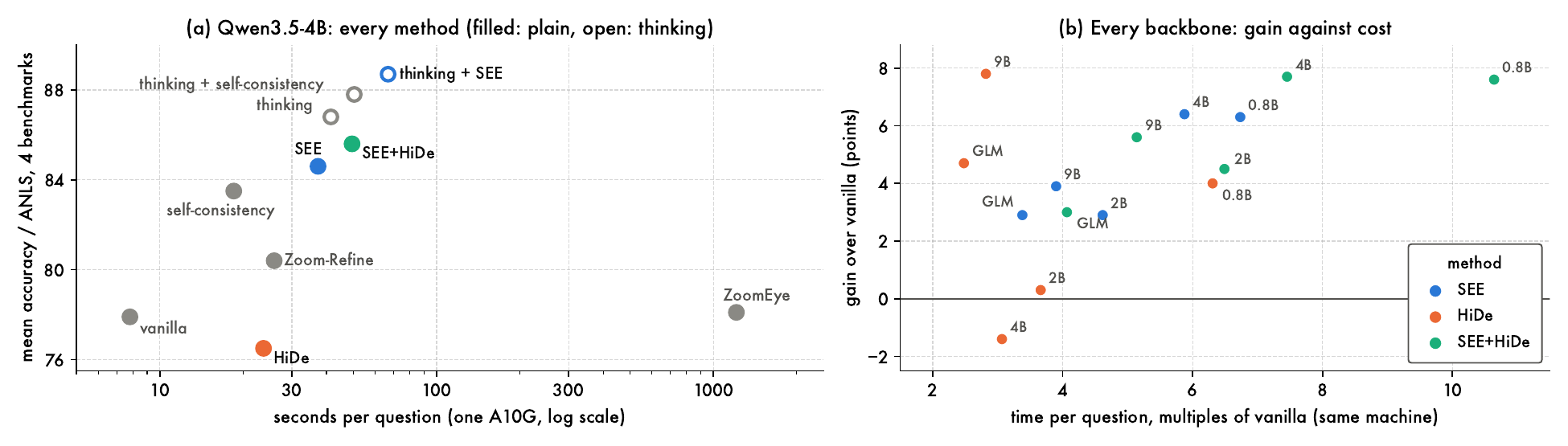}
\caption{\textbf{Cost against accuracy for every method}, with time per question as a multiple of the direct answer in the same
run. (a)~Qwen3.5-4B: mean over the four benchmarks (Table~2; thinking rows: Table~\ref{tab:supp:think}).
(b)~Every backbone: mean gain over the direct answer (Table~2).}
\label{fig:supp:cost}
\end{figure*}
\begin{figure*}[!t]
\centering
\includegraphics[width=0.86\textwidth]{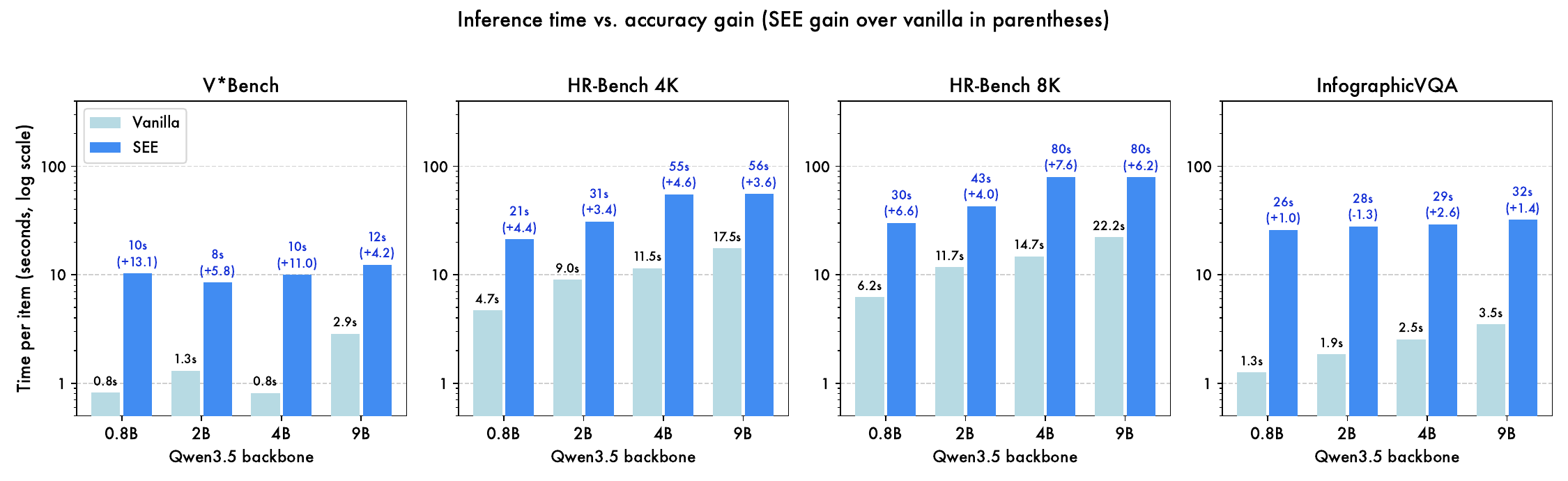}
\caption{\textbf{SEE's time per benchmark and Qwen3.5 size}: expected seconds per item for the direct answer and for SEE
(per-stage timings of the full runs, gate included), with SEE's gain in parentheses (ANLS on InfographicVQA). Each
size ran the direct answer and SEE on the same machine (one A10G for 0.8B--4B, except 4B V*Bench on one L40S; four
A10Gs for 9B), so only ratios within a size are comparable; HiDe's times are in Table~\ref{tab:supp:hide}.}
\label{fig:supp:latency}
\end{figure*}

\begin{table}[!t]
\centering
\caption{\textbf{HiDe, SEE, and SEE with HiDe's answer as one more view} (full sets; accuracy, ANLS on InfographicVQA; SEE:
analysis run at 4B; values on the questions both runs completed, so they can differ by $0.1$ from
Table~\ref{tab:full-results}). \emph{+HiDe}: SEE's gate and vote with HiDe's answer added as a view when the gate does
not stop, weighted by the confidence of its answer (rule fixed before scoring). $^\dagger$: not significantly
different from SEE ($p\ge0.05$; exact McNemar, sign-flip on ANLS). \emph{Either}: HiDe or SEE is right
(ANLS${\ge}0.5$). \emph{Time}: HiDe's seconds per question on A10G GPUs. GLM-4.1V InfographicVQA: $499$ questions.
Van.: direct answer. Bold: best of HiDe, SEE and +HiDe.}
\setlength{\tabcolsep}{3pt}
\resizebox{\columnwidth}{!}{%
\begin{tabular}{llcccccc}
\toprule
\textbf{Model} & \textbf{Bench.} & \textbf{Van.} & \textbf{HiDe} & \textbf{SEE} & \textbf{+HiDe} & \textbf{Either} & \textbf{Time} \\
\midrule
Q3.5-0.8B & V* & 66.0 & 74.3 & \textbf{79.1} & 78.5$^\dagger$ & 88.5 & 12.7 \\
Q3.5-0.8B & HR-4K & 66.1 & 70.8 & 70.5 & \textbf{72.1} & 81.6 & 27.4 \\
Q3.5-0.8B & HR-8K & 61.2 & 66.1 & 67.9 & \textbf{69.1}$^\dagger$ & 78.2 & 30.2 \\
Q3.5-0.8B & InfoVQA & 57.6 & 55.6 & 58.6 & \textbf{61.5} & 70.0 & 11.7 \\
\midrule
Q3.5-2B & V* & 75.9 & 73.8 & 81.7 & \textbf{83.8}$^\dagger$ & 91.6 & 10.5 \\
Q3.5-2B & HR-4K & 72.5 & 73.2 & 75.9 & \textbf{76.1}$^\dagger$ & 86.2 & 33.4 \\
Q3.5-2B & HR-8K & 69.8 & 71.2 & 73.8 & \textbf{75.0}$^\dagger$ & 84.1 & 37.9 \\
Q3.5-2B & InfoVQA & 72.8 & 74.2 & 71.5 & \textbf{74.5} & 81.0 & 5.7 \\
\midrule
Q3.5-4B & V* & 78.5 & 73.3 & 89.5 & \textbf{90.1}$^\dagger$ & 94.8 & 9.4 \\
Q3.5-4B & HR-4K & 77.6 & 76.0 & 82.2 & \textbf{84.6} & 92.6 & 32.2 \\
Q3.5-4B & HR-8K & 73.4 & 73.1 & 81.0 & \textbf{82.5}$^\dagger$ & 89.9 & 38.8 \\
Q3.5-4B & InfoVQA & 82.0 & 83.7 & 84.7 & \textbf{85.3}$^\dagger$ & 88.1 & 14.3 \\
\midrule
Q3.5-9B & V* & 85.9 & \textbf{92.7} & 90.1 & 92.1$^\dagger$ & 94.8 & 14.6 \\
Q3.5-9B & HR-4K & 76.8 & \textbf{87.4} & 80.4 & 82.8 & 91.1 & 48.0 \\
Q3.5-9B & HR-8K & 73.2 & \textbf{85.8} & 79.5 & 81.0$^\dagger$ & 91.1 & 57.4 \\
Q3.5-9B & InfoVQA & 85.9 & 86.9 & 87.2 & \textbf{88.2}$^\dagger$ & 91.5 & 9.9 \\
\midrule
GLM-4.1V & V* & 75.4 & \textbf{80.1} & 79.6 & 79.6$^\dagger$ & 86.9 & 8.1 \\
GLM-4.1V & HR-4K & 71.5 & \textbf{77.9} & 73.8 & 74.5$^\dagger$ & 80.9 & 12.5 \\
GLM-4.1V & HR-8K & 62.2 & \textbf{70.1} & 67.2 & 67.0$^\dagger$ & 76.4 & 13.4 \\
GLM-4.1V & InfoVQA & 83.7 & 83.5 & \textbf{83.8} & \textbf{83.8}$^\dagger$ & 88.1 & 7.9 \\
\bottomrule
\end{tabular}}
\label{tab:supp:hide}
\end{table}

\section{Baselines}
\label{sec:supp:baselines}
\subsection{Implementations}
All baselines run on the same frozen weights, image resolution limits and answer scoring as SEE, with their published
settings and nothing tuned on our data; they are re-implemented, or ported from the authors' code (HiDe~\cite{hide2025}, DyFo~\cite{dyfo2025}). A
baseline that fails on a question (out of GPU memory) returns the direct answer: Zoom-Refine~\cite{zoomrefine2025} on Qwen3.5-9B HR-Bench 4K
and 8K ($20$ and $90$ questions); the exception is ZoomEye~\cite{zoomeye2024} on Qwen3.5-9B InfographicVQA, whose $6$ failed questions are
left out ($494$ of $500$).
\emph{Self-consistency} (our adaptation of Wang et al.~\cite{selfconsistency2022}): the direct view plus three chains sampled with the evidence-seeking prompt of
App.~\ref{sec:supp:prompts}; each chain is answered as a view, and the answer is chosen by SEE's gate and confidence
vote, so the baseline matches SEE in everything except where its evidence comes from.
\emph{Zoom-Refine}: the model gives one box with its initial answer; the full image and the zoomed box, with the initial
answer in context, are then answered again.
\emph{ZoomEye}: the published Global+Local recipe with the release's constants (confidence from the Yes/No logits;
search thresholds $0.6$ and $0.8$, weight $0.6$, depth bonus $2$; a node is split while its longer side exceeds $448$
pixels). 
\emph{DyFo}: ported from the authors' release with its constants: $12$ Monte Carlo tree-search simulations, depth at most
$4$, exploration constant $1.0$, a focus and a $1.5\times$ zoom-out action, $20$-pixel padding, a reward of one minus the
crop's share of the image when the detector confirms every key object, and a reward-weighted vote over the answers at
all nodes with a full-image fallback. The release's visual expert keeps only GroundingDINO's boxes, so we call
GroundingDINO-base~\cite{groundingdino2024} directly with the release's thresholds (box $0.3$, text $0.25$).
\emph{HiDe}: Sec.~\ref{sec:supp:hide}.

\paragraph{Why the baselines break more right answers.} Zoom-Refine, ZoomEye and HiDe each make one final pass, so
whatever misleads that pass decides; DyFo votes over its search nodes but weights them by crop size and detector
confirmation, not by the model's confidence. On small objects they break up to $9.3\%$ (self-consistency), $16.4\%$
(Zoom-Refine), $18.9\%$ (HiDe) and $26.1\%$ (DyFo) of the right direct answers (Table~3). HiDe fixes more
errors than SEE, but at 4B it loses $23.7$ points on V*Bench relative-position questions, whose layout its re-packed image
drops (Sec.~\ref{sec:supp:hide}); DyFo's tight crops lose the context HR-Bench needs ($-5.9$ to $-12.9$ at 0.8B--4B;
Table~\ref{tab:full-results}).

\subsection{HiDe on every backbone}
\label{sec:supp:hide}
HiDe~\cite{hide2025} is run from its released code with the settings of its paper: attention layer $15$
(1-indexed), Gaussian $\sigma{=}3$, binarization threshold $0.7$, $256$ to $16{,}384$ visual tokens, greedy decoding;
the search prompt, the $8$ noise tokens, the prompts and the layout-preserving packing are taken verbatim from the
release. Nothing is tuned on our data. Layer $15$ of Qwen3.5 is a linear-attention layer, so its nearest softmax
layer, $16$, is used; GLM-4.1V uses layer $15$. HiDe makes $3$ model calls per question (entities, one attention pass,
the answer on the original and the packed image). Vanilla comes from the same runs and matches the main text. On
Qwen3.5-4B, HiDe's V*Bench accuracy is $84.3$ vs.\ $77.4$ (vanilla) on attribute questions and $56.6$ vs.\ $80.3$ on
relative-position questions.

\paragraph{Complementarity.} HiDe and SEE fail on different questions: per backbone, one of the two is right on
$79.6$--$92.1\%$ of the questions (InfographicVQA: ANLS $\ge 0.5$), against $69.0$--$88.2\%$ accuracy for the better of
the two alone. Adding HiDe's answer to SEE's vote
raises SEE on $16$ of the $20$ cells, by $+1.3$ (0.8B), $+1.6$ (2B), $+1.3$ (4B), $+1.7$ (9B) and $+0.1$ (GLM-4.1V)
on average, significantly on $5$ cells; it adds HiDe's three calls on the questions the gate does not stop. SEE+HiDe stays below HiDe alone at 9B and on GLM-4.1V on
V*Bench and HR-Bench: there SEE's gate answers $53$--$78\%$ of the questions directly, so HiDe is never consulted on them. The combination needs attention maps, so it is no longer black-box.

\section{Cost and inference time}
\label{sec:supp:latency}

Fig.~\ref{fig:supp:cost} puts every method on one cost axis, the time per question as a multiple of the direct answer in
the same run.
\paragraph{Compute in Fig.~1(c).} Compute per question counts every call SEE makes: $2N_{\text{LM}}$ FLOPs per input and output token plus $2N_{\text{ViT}}$ per vision patch of every image in the call. The image is re-encoded on every call and the gate is applied, so a confident question costs one call. Image tokens follow each model's own resizing; each \texttt{where} reply is counted at $30$ tokens, and the attention term that grows with sequence length is ignored. We average over each benchmark's questions, then over the four benchmarks. At 4B, SEE costs $3.65\times$ the FLOPs of the direct answer, less than its $5.9\times$ in time, because decoding is cheap in FLOPs but slow per token.

Fig.~\ref{fig:supp:latency} gives SEE's time per question for every benchmark and Qwen3.5 size. The multiple of the
direct answer varies with the share of questions the gate passes: it is highest on InfographicVQA for the small models,
whose direct answers are rarely confident, and lowest where the gate stops most questions.

\end{document}